# Open Problems in Frontier AI Risk Management

**Authors:**[*] Marta Ziosi,[1,] Miro Plueckebaum,[1] Stephen Casper,[2] Henry Papadatos,[10] Ze Shen Chin,[1,11] Peter Slattery,[3] James Gealy,[10] Tim G. J. Rudner,[1,6,9] Brian Tse,[13] Ariel Gil,[14] Patricia Paskov,[1] Maximilian Negele,[1] Rokas Gipiškis,[8,11] Nada Madkour,[16] Vera Lummis,[4] Rupal Jain,[7] Luise Eder,[1] Kristina Fort,[17] Malou C. van Draanen Glismann,[4] Inès Belhadj,[17] Amin Oueslati,[12] Anna K. Wisakanto,[15] Richard Mallah,[15] Koen Holtman,[11] Ranj Zuhdi,[17] Daniel S. Schiff,[5] Jessica Newman,[16] Malcolm Murray,[10] Robert Trager[1]

**Affiliations:** [1]Oxford Martin AI Governance Initiative, University of Oxford; [2]MIT Computer Science and Artificial Intelligence Laboratory, MIT; [3]MIT Future Tech; [4]Stanford University; [5]Governance and Responsible AI Lab, Purdue University; [6]University of Toronto; [7]Mercatus Center, George Mason University, [8]Vilnius University; [9]Vijil; [10]SaferAI; [11]AI Standards Lab; [12]The Future Society; [13]Concordia AI; [14]Pivotal Research; [15]Center for AI Risk Management & Alignment, [16]UC Berkeley Center for Long-Term Cybersecurity; [17]Independent

**Abstract**: Frontier AI systems - general-purpose systems capable of performing a wide range of tasks – bring a set of safety risks which risk management can help tackle. However, most AI-specific risk management standards were developed for narrow AI systems, before the advent of frontier AI. Frontier AI both amplifies existing risks and introduces qualitatively novel challenges. Not only is there a notable lack of stable scientific consensus resulting from the rapid pace of technological change, but emerging frontier AI safety practices are often misaligned with, or may undermine, established risk management frameworks. To address these challenges, we systematically surface open problems in frontier AI risk management. Adopting a problem-oriented approach, we examine each stage of the risk management process - risk planning, identification, analysis, evaluation, and mitigation - through a structured review of the literature, identifying unresolved challenges and the actors best positioned to address them. Recognising that different types of open problems call for different responses, we classify open problems according to whether they reflect (a) a lack of scientific or technical consensus, (b) misalignment with, or challenges to, established risk management frameworks, or (c) shortcomings in implementation despite apparent consensus and alignment. By mapping these open problems and identifying the actors best positioned to address them - including developers, deployers, regulators, standards bodies, researchers, and third-party evaluators - this work aims to clarify where progress is needed to enable robust and meaningful consensus on frontier AI risk management. The paper does not propose specific solutions; instead, it provides a problem-oriented, agenda-setting reference document, complemented by a living online repository, intended to support coordination, reduce duplication, and guide future research and governance efforts.

[*]Given this document's scope, inclusion as an author does not necessarily entail endorsement of all aspects of the report. Acknowledgements go to: Edward Kembery,[17] Zaina Siyed,[17] Francesca Gomez,[17] Ayrton San Joaquin,[11] Alexander Zacherl.[17]

# Introduction

Frontier AI poses significant safety risks (Bengio et al., 2026). It broadens access to tools for generating deceptive or harmful content (Achanta, 2025), exacerbates national security threats by enabling sophisticated offensive cyber capabilities (Moix et al., 2025; Potter et al., 2025), heightens inequalities through biased outputs (Gallegos et al., 2024), to cite a few. Traditionally, risk management offers a useful framework to identify, analyse and mitigate safety risks. Risk management processes operate at multiple levels: through high-level principles and processes for managing risks to organisations (e.g., ISO 31000:2018); through sector-specific standards for managing risks associated with particular classes of products (e.g., ISO 14971:2019 for medical devices); through guidance on selecting among relevant risk assessment techniques at different stages of the risk management process (e.g., IEC 31010:2019); and through overarching frameworks for integrating safety considerations across the risk management process (e.g., ISO/IEC Guide 51:2014).

In the context of AI, existing risk management standards primarily address narrow AI systems (e.g., ISO/IEC 23894:2023, ISO/IEC 42001:2023). These instruments were largely developed prior to the emergence of 'frontier' or 'general-purpose' AI: 'AI systems that learn patterns from large amounts of data, enabling them to perform a variety of tasks' (Bengio et al., 2026, p.17). This development both amplifies existing risks and introduces qualitatively novel challenges. Not only is there a notable lack of stable scientific consensus resulting from the rapid pace of technological change (Roberts & Ziosi, 2025); safety practices for frontier AI that are emerging are not fully aligned with, or may even undermine, established risk management processes (Koessler & Schuett, 2023; Schuett et al., 2023). Concurrently, improvements to and proposals for frontier AI risk management are being pursued along several distinct fronts. These include easily updatable, specific technical guidance (e.g., FMF, 2025; UK AISI, 2025), mapping the existing consensus on AI safety risks and practices (e.g., Bengio et al., 2026), independent proposals (e.g., (Barrett et al., 2025; Campos et al., 2025; Shanghai Artificial Intelligence Laboratory & Concordia AI, 2025) and regional efforts (Cyberspace Administration of China, 2025; EU Commission, 2025; NIST, 2024). Without a cohesive effort to systematically surface the challenges that frontier AI poses to risk management, however, these initiatives risk relying on flawed assumptions about the state of the field, they may fail to deliver targeted and meaningful progress, generate duplicative work, and create confusion or divergence over what should be applied in which contexts.

To address this, we propose to systematically surface open problems in the field of frontier AI risk management. We take a problem-oriented approach to advance the field by shedding light on what needs addressing, historically common in other disciplines (e.g., (Hilbert, 1900), and recently used to advance other emerging challenges in frontier AI (Barez et al., 2025; Casper, O'Brien, et al., 2025; Reuel et al., 2025; Sharkey et al., 2025). Our goal is twofold: 1) to highlight which challenges must be addressed such that meaningful and robust consensus on AI risk management can be pursued and 2) to pave the way for future solutions by formulating research questions and pinpointing which actors ought to pursue them. We do so by systematically examining each stage of the risk management process, conducting a review of the relevant literature (Grant & Booth, 2009) for each stage and identifying the 'open problems' and relevant actors to address them. Given that different kinds of open problems may require different approaches, we have classified the identified open problems according to whether they reflect (a) a lack of scientific (or technical) consensus, (b) misalignment with or challenges to established risk management frameworks, or (c) shortcomings in implementation or application despite consensus and alignment.

By 'open problems,' we refer to unresolved issues concerning the processes and techniques that organisations should implement to manage AI-related risks effectively. Accordingly, the paper does not focus a priori on a predefined set of risks from AI, but rather on the organisational and procedural mechanisms through which risks are identified, assessed, and mitigated. While the analysis primarily concerns strategies available to organisations developing, deploying and integrating AI systems, it also considers the roles of other relevant actors, such as regulators, academic researchers, standards developers, and third-party auditors, insofar as they are relevant to shape or support effective risk management processes. Additionally, the classification into different types of open problems can help inform which kinds of efforts are needed to address them. However, we refrained from proposing specific solutions as they may be best formulated by situated actors. The concrete outcome of this work is a problem-oriented, agenda-setting reference document, complemented by a [living repository hosted online](), intended to help relevant stakeholders identify gaps, coordinate action, and collectively advance better practices.

We recognise a few caveats and limitations. While our approach is systematic, the list of problems does not aim to be exhaustive, but at best illustrative of a range of relevant problems. Many of the open problems discussed arise precisely because there has already been substantial progress in these areas such that underlying challenges are becoming visible. Consequently, areas where we have identified relatively few open problems should not be understood as being more well developed or of lower importance; but instead as areas that remain insufficiently understood and explored such that we can clearly identify and articulate the relevant challenges. We hereby use the term 'problems' as a useful heuristic that should not be taken to describe issues that are inherently negative nor fully solvable, but that also includes persistent challenges that need to be constantly managed, productive disagreements or differing approaches with their own advantages and disadvantages. The aim of this work is therefore to surface and clarify such issues, rather than to claim their definitive resolution.

In order for this document to encourage alignment between traditional risk management and frontier AI risk management practices and frameworks, the structure of the paper will survey, as much as possible,[1] the open problems found following the high-level structure of existing risk management standards ([ISO 31000:2018](), [ISO/IEC 23894:2023]()), informed by safety-relevant standards ([ISO/IEC Guide 51:2014]()). The document is organised in the following sections: [1. Risk Planning](), [2. Risk Identification](), [3. Risk Analysis](), [4. Risk Evaluation](), and [5. Risk Mitigation](). We leave out transversal aspects such as Communication and Consultation, Monitoring and Review, and Recording and Reporting, also presented as 'risk governance' in other recent framework proposals, in order to keep the scope manageable. However, we may include them in further iterations.

## The Role of Risk Management for Frontier AI

Risk management encompasses the set of activities through which the likelihood of a risk occurring and the severity of its consequences is eliminated or reduced to an acceptable level (Vasvári, 2015). Although the specific structure and requirements of risk management processes vary across standards and protocols, several core stages can be identified at a high level (Vasvári, 2015). These typically include risk planning,[2] risk identification, risk analysis, risk evaluation, and risk mitigation (Figure 1):

1. **Risk planning (Section 1)** enables establishing the scope, context, and the objectives of the risk management process, as well as the criteria used to measure the significance and the acceptability of risk.
2. **Risk identification (Section 2)** surfaces risk sources, potential events, controls and consequences.
3. **Risk analysis (Section 3)** allows to gather information and conduct assessments to determine the consequences and likelihood of risk.
4. **Risk evaluation (Section 4)** helps determine the significance of risk with relevance to the pre-established criteria and make decisions on the acceptability of risk or its mitigation.
5. **Risk mitigation (Section 5)** involves risk reduction until acceptability is reached.

---

[1] The structure varies even across risk management standards, so the structure of the paper does not faithfully represent each and every existing risk management standard.

As mentioned above, most risk management standards also incorporate a set of transversal activities, such as recording and reporting, monitoring and review, and communication and consultation (e.g., IEC, 2019; ISO, 2018; ISO/IEC, 2023), which are not included in the following sections.

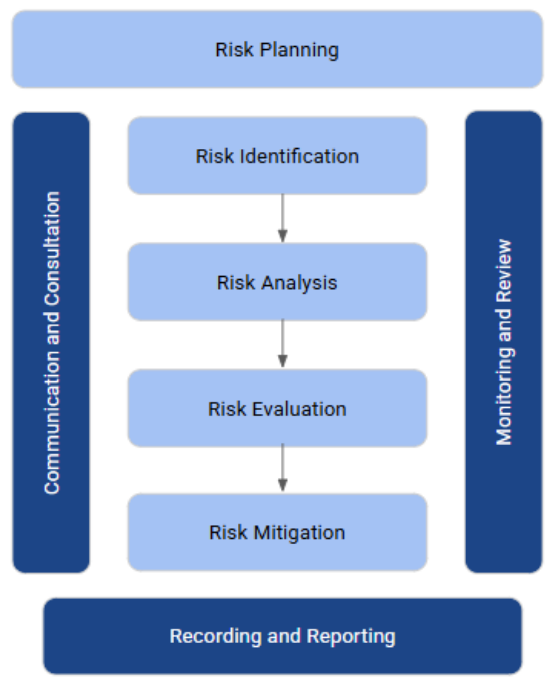

*Figure 1. Risk Management Process (adapted from ISO, 2018)*

Risk management provides a particularly useful analytical and practical lens through which safety risks can be addressed. Beyond reducing risks to acceptable levels, risk management supports a range of complementary organisational functions, including regulatory compliance, assurance, internal decision-making, and effective and efficient core organisational processes (Hopkin, 2010). It also ensures that key principles underpin these practices, including proportionality, alignment with organisational objectives, comprehensiveness, organisational embedding, and a dynamic and iterative attitude towards processes (Hopkin, 2010). From a product-safety perspective, it helps safely drive the system design and operation by iterative hazard and risk reduction (Preyssl, 1995), to prioritise the allocation of resources by ranking risk reduction efforts (Pidgeon, 1991), to support the tracking and verification of such efforts and prevent serious incidents (F. Khan et al., 2015).

---

[2] The proper name of this phase in risk management standards (e.g., ISO, 2018) is 'Establishing the scope, context, and criteria'. In this paper, we refer to this phase as 'Risk planning' for ease, but we mean the same thing.

There exists a well-established body of literature on risk management (Crockford, 1982; Dionne, 2013; Gahin, 1971; Hopkin, 2010; Vasvári, 2015; Williams & Heins, 1976), on system safety engineering (Bahr, 2015; Leveson, 2012), reliability engineering (Bergman, 1992) and hazard and risk analysis techniques (Ericson II, 2015). However, the application of these approaches to frontier AI remains comparatively limited. There is some emerging work that recognises the importance of applying risk management best practices to frontier AI. Wisakanto et al. (2025) propose systematic methodologies for risk identification and analysis, while Campos et al. (2025) propose an integrated risk management framework, both adapted from established practices in nuclear power and aviation. Analyses by Pouget & Zuhdi (2024) and Ziosi et al. (Ziosi et al., 2025) identify standards and guidance gaps by comparing frontier AI practices with mature standards in healthcare and other sectors. By surfacing open problems for each stage of the risk management lifecycle, this paper also aims to promote alignment between traditional risk management and emerging practices for frontier AI safety and thus further builds on this body of work.

Importantly, this paper focuses on frontier AI risk management from a safety perspective. We draw on established risk management standards, adopting the high-level structure of ISO 31000:2018 (ISO, 2018) and ISO/IEC 23894:2023 (ISO/IEC, 2023) (as per *Figure 1*), the selection of specific techniques from IEC 31010:2019 (IEC, 2019), and placing safety first by prioritising and integrating relevant elements of ISO/IEC Guide 51:2014 (ISO/IEC, 2014) throughout. Generic risk management standards such as ISO 31000:2018 provide a broad framework for managing risks across all organisational activities, where risk is defined broadly as the effect of uncertainty on objectives, encompassing both negative consequences and potential opportunities (IEC, 2019). In contrast, safety-oriented standards such as ISO/IEC Guide 51:2014 focus specifically on preventing or reducing harm to people, society and infrastructure by addressing hazards and reducing safety risks, defining risk in terms of the probability and severity of harm to people (ISO/IEC, 2014). While ISO 31000:2019 provides a useful high-level framework, we prioritise ISO/IEC Guide 51:2014 for content as it provides more targeted guidance for safety-critical contexts where protecting human life and physical integrity is the primary objective, such as frontier AI.[3]

Additionally, in order to bridge the gap between risk management and existing AI safety practices, we identify specific AI-relevant sub-sections for each main section. The separate sub-sections reflect a conceptual distinction rather than separate procedural steps, with some overlap to be expected in the use of specific techniques. For each sub-section, we outline: (1) what it entails, in safety risk management terms; (2) the current state of practice, including relevant standards; (3) the specific challenges introduced by frontier AI; and (4) outstanding open problems. Finally, all the risk management terms included here are defined in ISO 31073:2022 (ISO, 2022, freely available). A Glossary for AI terms can be found in ISO/IEC 22989:2022 (ISO, 2022b).

---

[3] Our safety-oriented focus is also reflected in our use of the term 'risk mitigation' to denote measures that reduce the likelihood or severity of harm, corresponding to ISO Guide 51's concept of 'risk reduction,' rather than the broader ISO 31000 category of 'risk treatment.' This choice signals that we exclude treatment options that do not themselves reduce harm, such as risk transfer or acceptance.

# 1. Risk Planning

Risk planning entails establishing the scope and context, setting objectives and the criteria for the risk management process (ISO, 2018). Safety risk management points to relevant aspects such as defining the system purpose and boundaries, its operational environment, the identification of likely users, the setting of safety goals (e.g., 'no single failure shall cause loss of life') and risk criteria (ISO/IEC, 2014). Overall, risk planning is a key step: it determines what kinds of impacts are emphasised, which stakeholders are engaged with, what tools and activities are employed, how success or failure is measured, and what criteria one adopts to determine how much risk is acceptable. Below, we review the open problems that frontier AI more specifically poses to each of these steps.

## 1.1 Establishing the Scope and Context

Establishing the scope and context involves defining the intended application of an AI system, including its functional domains, policy sectors of application, operating conditions, and system boundaries. It also requires specifying relevant aspects of the internal context, such as the AI models and data used, relevant features of the organisation, and the external context of deployment, including downstream uses and broader socio-economic or political factors. Below, we review some of the open problems related to this step.

**Intended (and unintended) use of the system.** Key in establishing the scope is making explicit the intended use of the system (ISO/IEC, 2014). This step is emphasised in multiple frameworks (e.g., ISO/IEC, 2023; NIST, 2023), with NIST's AI Risk Management Framework (RMF) asking organisations to specify why the system is being developed, the system-specific features and requirements, who the relevant stakeholders are, and what context-specific settings will matter (NIST, 2023). NIST's AI RMF also refers to 'use-case profiles' to create sector-specific (e.g., finance, healthcare) guidance where usage and context vary (NIST, 2023). In the case of frontier AI, mapping out intended uses will require capability- and modality-specific considerations (NIST, 2024). Additionally, for risk management activities to be sufficiently responsive to possible frontier AI risks, decision-makers must look beyond intended uses (Boine & Rolnick, 2023; L. Gailmard et al., 2025; Galinkin, 2022; Mylius, 2025). Scope consideration thus also requires determination of unintended uses as well as 'foreseeable misuses' of AI systems, which could lead to intended or unintended harms. Frameworks for anticipating misuse remain immature, particularly for frontier systems whose capabilities depend on context, prompting, tool integrations, and downstream system interactions (Anderljung et al., 2023; Bengio et al., 2024; Campos et al., 2025; Raman et al., 2025). A growing number of systematic taxonomies of risks and harms is available (Bagehorn et al., 2025a; Shelby et al., 2023; Slattery et al., 2025), with increasing attention to societal, user-centered, and technically-oriented risks, as well as taxonomies focused on individual contexts, like chatbots, mental health, or deepfakes (Bird et al., 2023; Coeckelbergh, 2025; R. Zhang et al., 2025). Additional methods exist to support consideration of adversarial uses (Kumar et al., 2019; Perez, Huang, et al., 2022; Tidjon & Khomh, 2022). While documentation practices such as Model Cards (Mitchell et al., 2019) and emerging System Cards (e.g., OpenAI, 2024) attempt to codify intended uses and limitations, adoption is uneven (Bommasani et al., 2025; de Laat, 2021), and most organisations do not systematically enumerate misuse scenarios.

**Boundary determination.** Determining boundaries entails clarifying what assets, processes, or activities are covered by the risk management framework, taking into account the internal and external context of the organisation. In principle, this boundary is also closely tied to the AI system operational boundaries. AI risk management recognises this, with standards like ISO 42001:2023 suggesting that the organisation determines its role also relative to the AI system, including roles such as AI providers (e.g., platform, product and service providers), AI producers (e.g., AI developers), AI customers (e.g., AI users), to cite a few (ISO/IEC, 2023). Frontier AI, however, blurs the boundaries between these roles, both with some actors being both providers and producers (e.g., Microsoft, Google) and with AI systems themselves being re-used (e.g., fine-tuned) across sectors, challenging the very distinction between users and producers. Limited guidance is provided on how to draw boundaries in practice for composite, vendor-integrated, or highly modular systems, as it is the case for frontier AI. Despite calls for greater attention to supply-chain dynamics (Balayn et al., 2025; Widder & Nafus, 2023) and proposals such as AI bills of materials (BSI & ACN, 2025), there is no standardised way to enumerate dependencies or interface responsibilities during scope-setting. This creates blind spots at integration points and weakens downstream accountability.

**Classification regimes as heuristics.** Classification regimes aim to facilitate setting the scope by grouping AI systems into categories. Classification efforts found in regulatory approaches (whether focused on parsing AI tools by sectoral context, such as in the UK policy context, or by risk levels, such as in the EU AI Act (Roberts et al., 2023)) can streamline placement of an AI system into a relevant bucket, fast-tracking scoping decisions about relevant uses, plausible unintended uses, boundaries, and so on. However, the utility of classification regimes depends on the quality and robustness of the underlying classification schema (Mökander et al., 2023). Poorly specified categories may lead to inappropriate scoping decisions, particularly if systems are misclassified as low risk or narrowly associated with a single regulatory domain. Classification schemes may also be vulnerable to strategic behaviour, as actors seek to game thresholds or definitions to reduce regulatory burden (Mökander et al., 2023; Tlaie, 2024; Veale & Borgesius, 2021). These challenges are amplified for frontier AI systems, which are general-purpose, and thus not easily captured in one category, and deployed across jurisdictions with emerging, divergent or absent regulatory approaches (Roberts & Ziosi, 2025). Overlapping AI-specific and sectoral regimes further complicate organisational scoping decisions, making the establishment of a single, coherent risk management approach contentious. While 'crosswalk' exercises can help organisations navigate regulatory fragmentation (NIST, 2023), the absence of more comprehensive harmonisation or guidance leaves significant uncertainty about how scope should ultimately be defined or constrained.

**Stakeholders and affected parties.** Beyond defining the intended purpose and deployment context, scope-setting involves identifying relevant stakeholders (Deshpande & Sharp, 2022). Current AI risk management standards emphasise broad stakeholder identification. ISO/IEC 23894:2023, for example, highlights customers, partners and third parties, suppliers, end users, regulators, civil society organisations, affected communities, and society at large (ISO/IEC, 2023). IEEE 7010-2020 requires identifying both directly and indirectly impacted groups (IEEE, 2020). In the example of autonomous vehicles, for example, the standard lists users, drivers, pedestrians, ridesharing companies, taxicab unions, urban planners, parking enforcers, safety inspectors, transportation regulators, and disability advocates, each with potentially unique impacts to consider (IEEE, 2020). Along these lines, the OECD's framework for the classification of AI systems (OECD, 2022) encourages not just looking at an AI model or associated outputs, but also understanding underlying data and input, economic context, and even considerations of 'people and planet.' Impacts can thus pertain not only to individuals, but also communities, social groups, and ecosystems. Stakeholder identification is thus closely tied with understanding AI systems' penetration within and beyond their immediate context.

In the case of Frontier AI, this task is specifically challenging as it can be unclear what the downstream implications of an AI system are, or what impacts will be borne by users or other stakeholders. Several frameworks and tools aim to support stakeholder-focused scoping through structured impact assessments. IEEE 7010-2020 encourages internal and external consideration of possible impacts and stakeholder engagement prior to setting objectives (IEEE, 2020), NIST AI RMF prescribes mapping impacts to individuals up to communities and society as a whole (NIST, 2023) and tools like HUDERIA encourage stakeholder-based AI impact assessment (Council of Europe, 2024). Yet, the dimensions for assessment remain open and underspecified (e.g., well-being, fairness, privacy, safety), creating confusion around which one to apply and their implementation in practice (Deshpande & Sharp, 2022). Finally, while many frameworks urge participatory scoping, methods for involving affected stakeholders systematically remain limited or underutilised (Young et al., 2024), despite evidence that stakeholder engagement shapes fairer and more context-sensitive problem formulations (Deng et al., 2025).

> **Open Problems**
> 1. **How can the inclusion of unintended uses and reasonably foreseeable misuses be effectively operationalised and incentivised in considering possible uses of frontier AI systems?**
>    *Who: Frontier AI developers' security teams, security experts from other relevant sectors (e.g., cybersecurity), independent researchers in AI safety and misuse*
>    **Type:** Shortcomings in implementation or application
>
> 2. **How can risk management boundaries be consistently determined and operationalised for frontier AI systems that are modular, reused across contexts, and embedded in complex supply chains?**
>    *Who: Regulators, AI developers' (downstream and upstream), supply-chain experts, standards bodies*
>    **Type:** Misalignment with or challenges to traditional risk management
>
> 3. **How can AI classification regimes be designed to remain robust against strategic misclassification and jurisdictional fragmentation?**
>    *Who: Standards developers, Conformity assessment bodies and AI assurance providers, intergovernmental bodies*
>    **Type:** Shortcomings in implementation or application
>
> 4. **How can organisations meaningfully engage affected stakeholders in the planning stage in ways that translate participatory scoping into concrete risk management actions?**
>    *Who: Frontier AI developers, Applied ethics and HCI researchers, Civil society organisations and community advocacy groups, regulators*
>    **Type:** Shortcomings in implementation or application

## 1.2 Setting Objectives

Closely connected to setting the scope and context is the determination of clear and actionable objectives, needed to underpin choices throughout AI risk management, both at the level of the organisation and of the AI system. Below, we review some of the open problems in setting objectives.

**Specifying objectives.** Setting objectives entails specifying what the risk management process is intended to achieve. In current standards, objectives are understood to be multiple and context-dependent, and may differ in tenor: some organisations emphasise regulatory compliance, while others prioritise prosocial or economic goals. Safety risk management emphasises the importance of setting safety goals, such as that no single failure shall cause loss of life (e.g., FAA, 2024). Organisational standards such as ISO/IEC 42001:2023 lists objectives ranging from accountability and transparency to ensuring sufficient AI expertise (ISO/IEC, 2023). Depending on their breadth and specificity, articulated objectives shape downstream risk management choices, as each objective can imply different priorities, trade-offs, and evaluation criteria (Alvarez, 2025; Schiff, 2025). Objectives may also differ in their status, with some being legally binding and others aspirational, and may be pursued through programmatic risk management activities or higher-level organisational decisions (Manning, 2017). For frontier AI, defining organisational objectives is particularly challenging given the multi-purpose nature of the technology and given that it might lead to breakthroughs (e.g., scientific discoveries) that might change an organisation's previously-set objectives (Renieris et al., 2024). Still, identifying a set of first principles is important to ensure adaptation to technological advancements and avoid approaching unexpected changes ad hoc (Renieris et al., 2024).

**Timing objectives**. Setting objectives also entails determining when objectives should be specified and revised over the lifecycle of an AI system, which can introduce its own challenges (Logan et al., 2021). Existing frameworks often treat scope identification and objective setting as early-stage planning activities (ISO/IEC, 2023), while simultaneously recommending continuous and iterative refinement through feedback loops. For example, the NIST AI RMF's Govern and Measure functions emphasise ongoing iteration (NIST, 2023), and IEEE 7010-2020 encourages multiple internal feedback loops throughout (IEEE, 2020). For frontier AI, this tension between early specification and continuous revision is particularly pronounced (NIST, 2023). The adaptable and updatable nature of frontier AI systems raises questions about when organisations can reasonably be confident that they have sufficient information to set objectives in the first place. Moreover, objective setting is conceptually and procedurally entangled with scope-setting and findings from risk identification and risk analysis, making it difficult to treat objectives as fixed inputs determined a priori.

**AI System-level objectives.** Beyond high-level objectives, setting objectives also entails specifying desired properties and behaviours of the AI system itself. Standards such as the NIST AI RMF MAP function emphasise eliciting system requirements (e.g., privacy) and making design decisions that account for socio-technical implications in addressing AI risks (NIST, 2023). In practice, developers of AI systems formulate detailed specifications for how models should behave, which guide the work of technical teams and data annotators when training, evaluating, and refining systems. In the case of frontier AI, this has raised questions over what set of human values AI systems should be aligned with (Korinek & Balwit, 2022; Lazar & Nelson, 2023) especially when value alignment may be skewed by either the often limited diversity of companies' teams relative to the world at large or, corporate profit motives (Abdulla & Chahal, 2023; Maslej et al., 2025; Singh et al., 2024). To mitigate these kinds of problems, some researchers have proposed paradigms for designing AI systems that balance competing views (Ali et al., 2025; N. A. Caputo, 2024; Kirk et al., 2023; Sorensen, Jiang, et al., 2024; Sorensen,

Moore, et al., 2024). Inspiration can also be taken from how other institutions develop normatively-accepted ways to choose what principles to align with (e.g., democracy) (Gabriel, 2020). However, other researchers have expressed concerns about 'pluralism washing,' arguing that technical approaches to pluralistic alignment fail to address deeper problems related to systematic biases and social power dynamics in the AI ecosystem (Birhane et al., 2022; Dobbe et al., 2021; Kalluri, 2020; Sloane et al., 2022).

---

**Open Problems**

1. **How can a coherent and actionable set of risk management objectives for frontier AI systems be specified when objectives are multiple, potentially conflicting, and subject to change as system capabilities and organisational goals evolve?**
   *Who:* *Frontier AI developers' safety and policy teams, Management science and organisational governance researchers from other relevant fields*
   Type: <u>Misalignment with or challenges to traditional risk management</u>

2. **How should objectives be staged and revised over time, given persistent uncertainty and the tight coupling between objectives and other later steps in the risk management lifecycle?**
   *Who:* *Frontier AI developers' safety and policy teams, Management science and organisational governance researchers from other relevant fields*
   Type: <u>Misalignment with or challenges to traditional risk management</u>

3. **How to formulate AI system objectives in a way which meaningfully minimises societal harm in a diverse set of deployment situations, while reducing perverse institutional incentives?**
   Who: *Frontier AI developers' safety and policy teams, Independent governance and oversight bodies, Researchers in ethics, political philosophy, STS and AI alignment*
   Type: <u>Shortcomings in implementation or application</u>

---

## 1.3 Setting Criteria

An important step before proceeding with the process of risk assessment is to set criteria for decisions which will be taken later on in the process. This includes criteria for deciding whether risk is acceptable, criteria for measuring risk, and criteria for deciding between options (e.g., deployment decisions). Below, we review open problems for these steps.

**Setting risk acceptance criteria.** Setting risk acceptance criteria entails defining how much risk is considered acceptable and on what basis such judgments are made (ISO, 2018; ISO/IEC, 2014). In established risk management standards, these criteria are typically grounded in explicit methods for determining tolerable risk levels. IEC 31010:2019 describes techniques such as Risk Bearing Capacity (RBC), which specifies how much risk an organisation can take without jeopardising its stability or long-term goals, and ALARP/ALARA principles, which require safety-related risks to be reduced 'As Low As Reasonably Practicable' or 'As Low As Reasonably Achievable' (IEC, 2019). Risk acceptance criteria in the safety-critical industry evaluate the extent of harm, expressed as 'severity x likelihood[4]', from potential accidents or losses on people, society, infrastructure, etc., within a specific intended context. For instance, as expressed by (Koessler et al., 2024), in the U.S. aviation industry the probability of 'failure conditions which would prevent continued safe flight and landing' should not exceed $1 \times 10^{-9}$ (one in a billion) per flight-hour (FAA, 1988). In the UK nuclear industry, the risk of death is 'unacceptable' above $1 \times 10^{-4}$ per plant-year and 'broadly acceptable' if below $1 \times 10^{-6}$ per plant-year (ONR, 2020).

In frontier AI, by contrast, risk acceptance criteria are most often operationalised through capability thresholds, defined as levels of system performance or capability that trigger specific mitigation measures (Campos et al., 2025). These thresholds use capabilities as a proxy for risk rather than expressing risk directly in terms of likelihood and severity of harm. While some frameworks link capability thresholds to generalised harm scenarios (Meta, 2025; OpenAI, 2025d), this approach primarily captures the possibility of harm and often neglects contextual and external factors, such as the threat landscape, that can influence both the scenario leading to the harm as well as risk levels (Caputo et al., 2025; Koessler et al., 2024). As a consequence, risk mitigation measures become harder to justify and calibrate because they are no longer directly tied to concrete accident or harm scenarios. To address these limitations, recent work offers guidance on the way capability thresholds can be operationalised more robustly, by using quantitative risk tiers and scenario-based risk modelling (Caputo et al., 2025; FMF, 2025b; Koessler et al., 2024; Murray et al., 2025; Wisakanto et al., 2025). Still, capability-centered criteria remain poorly suited for risks that are not revealed through model performance alone, such as risks to fundamental rights. While some propose harm modelling and associated thresholds for risks including toxicity, deception, discrimination and socioeconomic harms (Raman et al., 2025), others highlight that significant conceptual and practical challenges remain in translating such harms into risk management terms (Yeung, 2025). Others propose that these should be informed by input from the public (Choi & Rogers, 2025), such as via multistakeholder panels (Schuett et al., 2025).

**Conceptualising risks and establishing baselines.** Traditionally, safety-oriented risk management conceptualises risk in terms of the severity and likelihood of harm or consequences (ISO/IEC, 2014), and distinguishes between inherent risk (the level of risk before mitigations) and residual risk (the level remaining after mitigations). Most mature safety-critical industries rely on these concepts to evaluate whether systems meet predefined acceptability criteria within a stable baseline. In frontier AI, by contrast, many organisations increasingly rely on the concept of marginal risk, defined as the difference in risk relative to a chosen baseline (Williams et al., 2025). Similar approaches exist in other domains, such as European transport regulation, which uses the GAMAB principle ('Globalement Au Moins Aussi Bon') to require that new systems introduce no more risk than the existing state of the art (Tchiehe & Gauthier, 2017). Frontier AI developers, however, adopt heterogeneous baselines, including comparisons to human performance (K. L. Wei et al., 2025), earlier states of the world (Alaga & Chen, 2025; FMF, 2025c), a company's own previous model (AISI, 2025), or competitors' models (S. Williams et al., 2025). This widespread and inconsistent use of marginal risk introduces several challenges specific to frontier AI. Reliance on shifting baselines, particularly in a post–general-purpose-AI context, may enable a 'boiling frog' dynamic in which absolute risk increases across the ecosystem without being detected (Alaga & Chen, 2025). The absence of a shared baseline across organisations (FMF, 2025c), combined with competitive benchmarking against peers' models (Williams et al., 2025), may further amplify this effect while fostering a false sense of safety. There also remains the question of the marginal risk posed by open foundation models compared to closed models and non-AI sources for which Kapoor et al. (Kapoor et al., 2024) present an initial risk assessment framework.

---

[4] In risk management, likelihood is treated as synonymous with probability (ISO, 2022a, 3.3.16). This is in contrast to the usage in statistics, where likelihood refers to how likely a model has certain parameters given a certain outcome, while probability refers to how probable an outcome is given a certain set of model parameters. In this document, we only reference the notion of probability in statistics, but use the both terms 'likelihood' and 'probability' interchangeably.

**Criteria to decide between multiple options.** Criteria for deciding between multiple options determine how organisations choose among alternative courses of action (IEC, 2019), such as deployment, delay, restriction, or additional mitigation going beyond or, where relevant, complementing risk acceptance criteria. In practice, organisations often face decisions in which multiple, competing objectives are at stake and both potential harms and benefits must be weighed. Risk management standards describe several decision-support techniques for such contexts. Cost–benefit analysis (CBA) is a prominent approach, evaluating options based on expected financial or utility losses and gains (IEC, 2019). Other techniques include decision-tree analysis, which represents the utility of decisions in a structured, sequential form (Kirkwood, 2002), and multi-criteria decision analysis, which allows multiple criteria to be weighted and compared simultaneously (Velasquez & Hester, 2013). Under conditions of deep uncertainty, methods such as Robust Decision Making (RDM), widely used in climate and disaster risk assessment, stress-test candidate options across many plausible futures and evaluate them against multiple success metrics (Dittrich et al., 2016). For frontier AI, however, the suitability of these techniques remains largely untested and highly context-dependent. Some frontier AI frameworks make limited reference to cost–benefit considerations; for example, by pairing risk assessments with benefit assessments (Meta, 2025) or explicitly weighing risks and benefits when defining deployment standards (Anthropic, 2025). Yet CBA is poorly suited to safety-critical, high-uncertainty environments and struggles to account for public-good impacts and non-quantifiable harms (IEC, 2019), which are central concerns in frontier AI governance.

> **Open Problems**
> 1. **How can risk acceptance criteria for frontier AI systems be defined in ways that are meaningfully tied to harms?**
>    *Who: Frontier AI developers' safety and policy teams, Standards bodies, researchers in risk governance, ethics, and safety engineering, Affected Stakeholders from the public*
>    **Type:** Misalignment with or challenges to traditional risk management
>
> 2. **How can risk in frontier AI be conceptualised in a way that remains comparable and sensitive to ecosystem-level risk accumulation, given the coexistence of multiple approaches to risk measurement (e.g., marginal risk)?**
>    *Who: Frontier AI developers' safety and policy teams, AI safety researchers and third-party evaluators, Standards bodies, Intergovernmental bodies*
>    **Type:** Shortcomings in implementation or application
>
> 3. **Which decision-making approaches can support trade-offs between competing objectives in frontier AI decisions (e.g., deployment decisions) when uncertainty is high and many impacts are hard to quantify?**
>    *Who: Frontier AI developers' safety and policy teams, Researchers in decision theory and applied ethics, safety and organisational risk management experts*
>    **Type:** Misalignment with or challenges to traditional risk management

# 2. Risk Identification

Risk identification is the systematic process of discovering and cataloguing potential risks associated with an AI system, formally defined as the 'process of finding, recognising and describing risks,' which 'involves the identification of risk sources, events, their causes and their potential consequences' (ISO, 2022a, 3.3.9). Safety risk management places a stronger emphasis on 'hazards' as relevant 'sources of potential harm' (ISO, 2022a, 3.3.12), focusing on identifying hazards, as well as reasonably foreseeable hazardous situations and events (ISO/IEC, 2014). Risks can be identified through the use of existing documented risks, such as through taxonomies or repositories, or in a more open-ended manner such as through elicitation from stakeholders and experts (IEC, 2019). At the end of the risk identification process, a common practice is for the identified risks to be recorded in a risk register (Balfe et al., 2014). To support traceability and auditability, in addition to listing risks, the risk register should document the scope and assumptions under which risks were identified, the methods of identification, and versioned changes over time. The rest of the section is organised around the identification of risk sources (including hazards), potential events and outcomes, controls, and consequences. In practice, however, most of the risk-identification methods cited below generate information about several of these items simultaneously, and the boundaries between them frequently overlap. In fact, many of these techniques straddle across other risk management processes as well, such as risk analysis, risk evaluation, and risk mitigation. Below, we review open problems.

## 2.1 Identifying Risk Sources

A key aspect in the process of risk identification is to identify risk sources related to the development and use of frontier AI. This should be informed by and in line with the defined scope and context ([Section 1.1](#)), as well as objectives ([Section 1.2](#)) presented above. Below, we review open problems related to the steps involved.

**Considering AI risk sources.** Considering AI risk sources entails identifying the elements that can give rise to risk within an AI system and its operating context. In risk management, a risk source is defined as an element that alone or in combination has the potential to give rise to risk (ISO, 2022a, 3.3.10). Hazards constitute a specific class of risk source characterised by an inherent property, a condition or a state that can cause harm if realised or activated (ISO, 2022a, 3.3.12), such as high voltage or pathogenic agents. Other risk sources do not possess inherently harmful properties but generate risk through structural, behavioural, informational, or organisational characteristics, such as ambiguous decision authority, poor data quality, mis-specified system objectives. AI-related standards identify a wide range of such sources, including environmental complexity, lifecycle and hardware issues, lack of transparency, and technological readiness for a given application context (ISO/IEC, 2023, Annex B). Frontier AI systems introduce additional and distinct risk sources. Regulatory frameworks on frontier AI increasingly emphasise model capabilities (e.g., offensive cyber capabilities), model propensities (e.g., hallucination), model affordances and so-called 'other systemic risk sources' (e.g., model configurations, model properties and context) (EU Commission, 2025). The NIST Generative AI Profile further highlights human behaviour and human–AI interaction as critical sources of risk, including misuse, abuse, and unsafe repurposing (NIST, 2024). Complementing these approaches, repositories of AI-related threats such as the MITRE ATLAS Matrix catalogue AI-specific attack tactics across the system lifecycle (MITRE, n.d.), enabling identification of actors, assets, and conditions that may give rise to harm. Notwithstanding these resources, however, many frontier AI companies' safety frameworks focus predominantly on model capabilities, such as CBRN and AI R&D or 'AI self-improvement' (Anthropic, 2025; OpenAI, 2025d) and propensities, such as sandbagging or undermining safeguards (OpenAI, 2025d), with less systematic attention to affordances and deployment context as risk sources. This can obscure how configuration choices, access modalities, or integration pathways shape real-world risk, which is a key aspect of risk identification.

**Selecting techniques for identifying AI risk sources.** Selecting techniques for identifying AI risk sources entails choosing structured methods to surface relevant sources of risk across a system's design, operation, and context. Classical risk management standards and documents describe several such techniques (Ericson II, 2015; IEC, 2019). Hazard Identification (HAZID) is a structured brainstorming technique, typically applied early in a project, that aims to identify hazards, but also aims to elicit potential initiating events, high-level consequences, and existing or proposed controls to provide contextual information for subsequent risk assessment (Golwalkar & Kumar, 2022). Hazard and Operability Studies (HAZOP) systematically examine a system or process by identifying potential deviations from design intent and their possible causes and consequences using predefined guidewords (e.g. 'no', 'more', 'less') applied to various parameters related to the system (IEC, 2019; Mocellin et al., 2022). The Structured What-If Technique (SWIFT) similarly relies on guided brainstorming, using a predefined set of guidewords (e.g. timing, amount, etc.), combined with 'what if' questions to identify known risks, risk sources, and existing or proposed controls (IEC, 2019; Potts et al., 2014). Complementary backward-chaining techniques where the sources are identified through the events or the consequences, include Ishikawa (fishbone) analysis, which works backward from an identified event to surface possible causes by depicting the causes as the 'bones' of a 'fish' with the 'head' as the event (IEC, 2019), and Failure Mode and Effects Analysis (FMEA), which decomposes a system into elements and examines their failure modes, causes, effects, and associated controls (IEC, 2019).

For frontier AI, applying these techniques presents distinct challenges. The complexity, general-purpose nature, and non-linear interactions characteristic of frontier systems can strain methods originally developed for bounded, deterministic systems. Recent work proposes adaptations such as aspect-oriented hazard analysis, using first-principles taxonomies of AI system aspects, such as capabilities, domain knowledge, and affordances, to identify critical risks through the system's characteristics, context, and guided risk pathway threat modelling (Wisakanto et al., 2025). Koessler and Schuett (2023) cite the fishbone (Ishikawa) method as useful for identifying frontier AI risk sources in highly uncertain contexts by working backward from possible consequences, while cautioning that it is ill-suited for risks involving non-linear interactions such as competitive dynamics. They also recommend the use of risk taxonomies to reduce blind spots and foster a shared understanding of the risk landscape across stakeholders (Koessler & Schuett, 2023). Currently, there are several taxonomies or repositories of AI risks, such as AI Risk Categorisation Decoded (Y. Zeng et al., 2024), the AI Risk Repository (Slattery et al., 2025), OWASP Top 10 Risk & Mitigation (OWASP, 2025), and AI Risk Atlas (Bagehorn et al., 2025b). However, they also note that taxonomies are time-consuming to develop and may convey a misleading sense of completeness in the absence of empirical data (Koessler & Schuett, 2023).

> **Open Problems**
> 1. How can risk identification be systematically made to account for model affordances, deployment configurations, and human–AI interactions, beyond focusing on model capabilities and propensities?
> *Who:* *Frontier AI developers' safety teams, Third-party evaluators, Researchers in AI risk management and sociotechnical systems, Regulators*
> **Type:** <u>Shortcomings in implementation or application</u>

> **2. How can risk source identification for frontier AI systems account for complex interactions, non-linear dynamics and unknown risks that are poorly captured by existing techniques?**
> *Who: Frontier AI developers' technical safety and risk modelling teams, Researchers in complex systems and risk modelling, Third-party evaluators*
> Type: <u>Misalignment with or challenges to traditional risk management</u>

## 2.2 Identifying Potential Events, Controls and Consequences

Beyond, and sometimes alongside, identifying risk sources, it is key that one also identifies relevant elements like events, controls and consequences. We review open problems for each in turn below.

**Identifying potential events.** In risk management, an event is defined as an occurrence or change in a particular set of circumstances, and may have multiple causes and consequences (ISO, 2022a, 3.3.11). Classical techniques such as HAZID, HAZOP, and SWIFT support this step by moving beyond hazard identification to outline credible events that hazards may lead to, mapping initiating conditions to system deviations, failures, or losses (Ericson II, 2015; IEC, 2019) . Risk management standards for AI systems further recommend drawing on sources such as market data or incident reports on similar systems, usability studies, and interviews and reports from internal and external experts to inform event identification (ISO/IEC, 2023). For frontier AI systems, identifying potential events is particularly challenging due to limited historical experience and rapidly changing deployment contexts. There are a set of AI incident databases which can be useful in this respect. These include the AI Incident Database (AIID, n.d.) and the AI Incidents and Hazards Monitor (OECD, n.d.). These historical incidents serve as a reference for potential events should the identified risks not be appropriately managed. Nevertheless, as many of these databases are built from voluntary incident reports, they are not representative of the full range of potential events, as the events reported may be those that receive the most public attention instead of having the highest probability or severity. Furthermore, just as past performance in financial markets may not be indicative of future performance (Kahn & Rudd, 2019), historical AI incidents will necessarily not be reflective of future incidents that have yet to occur.

**Identifying controls.** Controls are defined as measures that maintain and/or modify risk (ISO, 2022a, 3.3.33). In the safety sense, this term is understood as focused on hazard elimination and risk reduction (risk mitigation) (ISO/IEC, 2014), and it identifies and documents controls relevant to understanding the risk. Classical risk management techniques such as HAZID, HAZOP, SWIFT, and FMEA are used to identify controls as part of its process. As failure modes and causal pathways are identified, these methods also outline planned controls that are part of the system's design to mitigate these pathways to harm. In the context of frontier AI risk management, there are also existing repositories of AI-related controls such as the MIT Risk Mitigation Taxonomy (MIT, n.d.-b), the Secure AI Framework (Google, n.d.), and other academic sources (Gipiškis et al., 2024a). Beyond merely listing controls, risk management standards also require that their operating effectiveness be taken into account, including the possibility of control failures (ISO/IEC, 2023). In traditional risk management, stating the effectiveness of controls is relatively tractable, as controls are often well-established with their limitations well-understood by practitioners through accumulated operational experience. For frontier AI, however, the evidence base for control effectiveness is severely limited. Many controls, such as those described in 5. Risk Mitigation, have been developed only recently, and some lack any sustained track record of deployment in real-world conditions. Assessing their operating effectiveness therefore involves substantial uncertainty, which means that controls identified for frontier AI systems often cannot be documented with the same confidence in their reliability as controls in more established domains.

**Identifying consequences.** Consequences are outcomes of an event affecting objectives (ISO, 2022a, 3.3.18), where the objectives would be set in the previous phase as discussed in Section 1.2 Setting Objectives. Classical risk management techniques for consequence identification closely overlap with those used for hazard and event identification, since identification of hazards naturally yields their associated events and consequences. Additionally, scenario analysis, which covers a range of techniques that involve developing models of how the future might turn out, can also be used as a forward-chaining technique to identify consequences (IEC, 2019). This can include extrapolating past trends for the short-term and building imaginary but credible scenarios for the long-term (IEC, 2019). This technique is also useful to inform risk analysis techniques later on, such as risk modelling (Section 3.3). For frontier AI systems, identifying consequences is particularly challenging due to the fast-pace of technological change and limited empirical understanding of real-world impacts. Recent efforts, such as AI-related scenario exercises developed by the Department for Science, Innovation and Technology in the UK (DSIT, 2024a), illustrate a wide spectrum of possible social, economic, technological, and environmental consequences, ranging from incremental adoption effects to catastrophic global outcomes. However, such scenario analyses are subject to significant uncertainty, which is amplified by the fact that frontier models are often assessed early in their lifecycle, while many consequential harms only emerge downstream when models are integrated into applications and deployed at scale (Touzet et al., 2025).

> **Open Problems**
> 1. **How can potential risk events for frontier AI systems be identified systematically, including events with no historical precedent, given the limitations (e.g., lack of representativeness) of incident databases and past-case evidence?**
>    *Who: Frontier AI developers' safety teams, AI incident database maintainers. security and incidents experts, AI safety and foresight researchers, intergovernmental bodies and regulators*
>    **Type:** Misalignment with or challenges to traditional risk management
>
> 2. **How can the operating effectiveness of controls for frontier AI systems be adequately characterised and taken into account during risk identification, given that many such controls lack sustained deployment evidence?**
>    **Who:** *Frontier AI developers' safety teams, Risk management and safety engineering researchers from other relevant sectors, Third-party evaluators*
>    **Type:** Lack of Consensus
>
> 3. **How can consequences of frontier AI systems be anticipated early in the lifecycle, when many harms emerge only downstream through integration into applications and broader sociotechnical systems?**
>    *Who: Frontier AI developers' safety teams, AI safety and Sociotechnical Researchers, Downstream Deployers, Users and other Affected Stakeholders*

> **Type:** Misalignment with or challenges to traditional risk management

# 3. Risk Analysis

Risk analysis generally refers to the process aimed at further comprehending the nature of risk and its characteristics (e.g., sources, events, consequences, etc. as identified in Section 2) and to determine the level of risk (ISO, 2022a, 3.3.15). Safety-focused risk management is specific about getting to an estimation of risk, assessing it in terms of its likelihood and severity, and with the precise aim to eventually inform an evaluation of its acceptability (ISO/IEC, 2014). It is also important to mention that there can be some overlap between techniques used in risk identification and analysis. Standards on risk assessment techniques such as IEC 31010:2019 catalogue a wide range of techniques that are used across both, including HAZOP and FMEA as cited above, as well as fault tree analysis, event tree analysis, and bow-tie analysis (IEC, 2019). For the purpose of this work, we include here any step that is aimed at further understanding the nature of risk and its characteristics, and assessing its severity and the likelihood of consequences. We leave more theoretical techniques that help us to analytically map out the risk space to Section 2. In what follows, we distinguish three stages of risk analysis: 1) Internal information gathering which is based on information available to model developers themselves through internal testing, 2) External information gathering which is based on data made available to developers based through third-party testing or external usage, and 3) Severity of the consequences and likelihood, expressing the level of risk. The division in these categories is compatible, yet slightly adapted from risk management in order to bridge the gap with relevant risk assessment methods in frontier AI. Below, we review open problems for each of these steps.

## 3.1 Internal Information Gathering

In order to further comprehend the nature of risk, we here review approaches that contribute to gathering relevant information about the nature of risk and its relevant characteristics referred to in Section 2 (e.g., risk sources or hazards, etc.) as available to model developers through internal sources and methods.

**Assessing an AI system's properties.** Assessing an AI system's properties entails analysing internal characteristics of the system that are relevant to understanding how identified risk sources, events, and impacts may arise. In risk management, this step typically involves gathering information about the type and significance of risk sources and analysing potential consequences through methods such as impact assessments, including assessments of the intended effects of AI development or use on individuals or society (ISO/IEC, 2023). These analyses provide internal information to eventually determine the severity and likelihood of risk (Section 3.3). In frontier AI practice, this step is predominantly operationalised through capability assessments or 'model evaluations.' Frameworks such as the NIST Generative AI Profile (2024) and the EU General-Purpose AI Code of Practice (2025) emphasise evaluations as a primary means of analysing model or system properties. Capability assessments are currently used to determine whether capability thresholds have been crossed and safeguards should be implemented (Anthropic, 2025; Google, 2025; OpenAI, 2025d); whether the system possesses vulnerabilities that could enable harmful misuse (Anil et al., 2024; Carlini et al., 2024; Sharma, Tong, Mu, et al., 2025); or whether the underlying model demonstrates undesirable propensities, values, bias or discriminatory tendencies (Greenblatt, Denison, et al., 2024a; S. Huang et al., 2025; Meinke et al., 2025; Solaiman et al., 2024). Evaluation results are thus used to guide a broad range of risk-management-relevant decisions such as deploying technical mitigations, restricting access, delaying deployment, or pursuing further investigation. Capability assessments, however, focus on what models can do, rather than directly measuring the risk that it poses (Koessler et al., 2024). As a consequence, caution is needed when interpreting and applying evaluation results for risk management as they alone cannot express risk. Additionally, even when applied correctly, there is still limited guidance on how evaluation outputs should be integrated into broader risk models in order for them to correctly inform a determination of risk (C. Yu et al., 2026).

**Ensuring quality, coverage, and robustness.** Ensuring quality, coverage, and robustness entails establishing that assessments used to analyse AI system properties are methodologically sound, sufficiently comprehensive, and reliable inputs to risk management decisions. While such assessments have improved in quantity and quality over the last few years, there remains a significant shortage of widely adopted and sufficiently high quality ones, and best practices are still evolving (Apollo, 2024). Many existing assessments vary substantially in their methodological rigor, scope, and reproducibility (Paskov et al., 2025; Reuel et al., 2024). Frontier AI systems amplify these challenges. Assessments might not elicit or represent a system's full capabilities. For instance, evaluations might assess models with less access to inference time compute, fewer attempts, or less effective tools and scaffolding than future deployments of the system might realistically have access to (Barnett & Thiergart, 2024; Götting et al., 2025; Turtayev et al., 2025). It is also challenging to develop capability assessments that are robust for systems with increasing capabilities (McKee-Reid et al., 2024; Summerfield et al., 2025; Von Arx et al., 2025), given that these might undermine assessments' accuracy (Greenblatt, Denison, et al., 2024b), particularly in more agentic scenarios (Anthropic, 2025). Assessments' results may also be highly sensitive to small differences in prompting and implementation (Biderman et al., 2024; Burden, 2024; Robinson & Burden, 2025; Schaeffer et al., 2023; Sun et al., 2025), thereby hampering their reliability and reproducibility. Taken together, these issues complicate comparisons across models and over time, limit confidence in reported results, and weaken the role of evaluations as stable inputs into downstream risk analysis. This is made even more difficult by a lack of detail and comprehensiveness in how results of safety evaluations are reported publicly (McCaslin et al., 2025; Paskov et al., 2025; K. L. Wei et al., 2025).

**Linking results to real-world behaviour and harms.** Linking evaluation results to real-world behaviour and harms entails assessing whether measured model capabilities and propensities reliably predict how AI systems will perform and affect outcomes in real-world settings. While current research has made progress, building robust predictive models of AI system capabilities (Hofstätter et al., 2025; L. Zhou et al., 2025), and how these might map to real-world scenarios and harms (Barnett & Thiergart, 2024; Mukobi, 2024) remains difficult, even more so for frontier AI whose behaviour and real-world impacts are even less predictable. Real-world tasks often involve fuzzy objectives and hard-to-measure outcomes. Increasingly with frontier AI, these might entail complex and interactive environments, which evaluations struggle to approximate (Phan et al., 2025) thus limiting their external validity, e.g., their ability to support inference about real-world impacts. Moreover, most current assessments focus on isolated models rather than interactive or multi-agent environments, which remain underdeveloped (AI Village, n.d.; Hammond et al., 2025), and evaluations involving representative numbers of human participants are uncommon beyond basic red-teaming exercises. Frontier AI company assessments also often under investigate the effects of directly uplifting human capabilities, including in crucial, high-risk domains such as cyber offense (Righetti, 2024). Finally, more realistic evaluations are frequently slow and costly, especially when they require large numbers of qualified participants or substantial compute and, particularly in national-security-relevant domains, may pose security risks.

> **Open Problems**
> 1. **To what extent can capability assessments be relied upon to inform risk-relevant decisions for frontier AI, and how should their results be integrated into broader risk models?**
>    *Who:* *Frontier AI developers' evaluation and safety teams, AI safety and evaluation researchers, Third-party evaluators, AI users and impacted stakeholders*
>    **Type:** <u>Misalignment with or challenges to traditional risk management</u>
>
> 2. **How can capability assessments be designed to remain valid and reproducible when evaluating fast-evolving frontier AI systems with increasing capabilities?**
>    *Who:* *Frontier AI developers' safety and evaluation teams, Third-party evaluators, AI evaluation researchers and benchmark developers*
>    **Type:** <u>Lack of Consensus</u>
>
> 3. **How can the ability of capability assessments to draw inferences about real-world impacts be improved, or appropriately caveated, particularly in complex multi-agent settings where impacts emerge beyond the model level?**
>    *Who*: *Frontier AI developers' safety and evaluation teams, Researchers in AI evaluation, risk modelling, and sociotechnical systems, AI users (e.g., Downstream deployers and affected stakeholders), Public-interest and policy-oriented research organisations*
>    **Type:** <u>Lack of Consensus</u>

## 3.2 External Information Gathering

Beyond gathering internal information, it is also important to engage with external actors or implement mechanisms to gather external information to inform a better understanding of identified risks. This is not only important to gather more information, but also to indirectly validate or stress-test the information gathered so far (e.g., capability assessment results), a key step in risk management (IEC, 2019). This step can also help identify new risks, trigger another round of risk assessments or revise the plan for risk management.

**External assessments.** External assessments involve independent actors to verify, validate, or scrutinise an organisation's internal risk assessment. Risk management standards reference such engagement at a high level, for example through communication with or participation of external stakeholders (ISO, 2018; ISO/IEC, 2023), and ISO/IEC 31010 explicitly calls for 'independent review processes' to verify and validate risk analysis (IEC, 2019, 6.4.1). Regarding frontier AI, both NIST AI RMF Generative AI Profile (2024) and the EU GPAI Code of Practice (2025) require the involvement of independent external assessors under specified conditions. These assessments, often referred to as external model or system evaluations (Sharkey et al., 2024; Xia et al., 2024) or technical audits (Kluge, 2023), typically focus on technical functionality, performance, or specific risk domains such as bias or chemical, biological, radiological, and nuclear (CBRN) risks (Brundage et al., 2026). While external evaluations are increasingly common in frontier AI risk management, there is little consensus on appropriate protocols, the scope of assessment, or how results should be disclosed and acted upon (Cattell et al., 2025; Lam et al., 2024). A central difficulty concerns the depth of access granted to external assessors (Homewood et al., 2025; Kembery & Reed, 2024). Proposals for 'structured access controls' suggest calibrating the level of access to the risk, enforced through technical controls ranging from API sampling to secure enclaves that enable parameter-level verification (Bucknall & Trager, 2023; Shevlane, 2022). However, it remains unclear how to strike a balance between meaningful access and preventing misuse, theft, or disclosure of sensitive materials. Black-box or output-only audits are insufficient for rigorous safety assessment (Casper et al., 2024), with studies suggesting that such testing under-detects risks (Che et al., 2025), while deeper forms of access (to training data and deployment information, up to the system's inner workings or internal model representations) may be necessary to reliably interpret model behaviour (Casper et al., 2024) yet may raise security concerns for developers. Importantly, the resulting problem of 'mutual privacy', protecting both proprietary designs and the results of independent testers, remains broadly unexplored despite proposals for secure technical environments, encrypted test setups, and contractual safeguards (Bucknall et al., 2025).

**Ensuring structural independence.** Ensuring structural independence entails designing external evaluation processes so that assessors can scrutinise AI systems with sufficient independence from developers. Risk management standards generally refer to independent review at a high level (IEC, 2019, 6.4.1, as above) but provide limited guidance on how to secure genuine independence in practice. Studies of early audit practices show that many assessments described as 'external' were in fact hybrid forms, in which the developer selects and finances the assessor, defines the scope, and may be able to veto the publication of negative results (Raji et al., 2022). Becerra Sandoval & Jing (2025) point out that these conditions can also shape methodology and timelines, favouring faster, scalable, quantitative techniques over slower socio-technical investigations, echoing well-documented conflicts of interest in financial auditing (Moore et al., 2006). Frontier AI heightens these concerns because evaluations are costly, technically complex, and often dependent on privileged system access. Proposals to mitigate capture dynamics include independent audit committees, public declarations of conflicts of interest (Lam et al., 2024), public or pooled funding of audits, regulator-managed auditor lists with rotating assignments to reduce familiarity bias, and a legal right to publish critical findings (Raji et al., 2022). In the context of AI evaluations specifically, some have proposed legal and technical safe-harbour agreements to protect good-faith evaluators from legal retaliation or the threat of account suspension (Longpre et al., 2024). However, operationalising such safeguards raises unresolved questions about enforcement, incentives, and governance. Furthermore, the lack of specialised expertise, computational infrastructure, and reproducible methods means that the ability to conduct rigorous external evaluations remains concentrated in a handful of well-resourced organisations, raising further concerns about the independence and diversity of oversight (Anderljung et al., 2023; Busuioc, 2022; Costanza-Chock et al., 2022).

**Post-deployment assessments.** Post-deployment assessments entail monitoring and reviewing systems after deployment to compare real-world outcomes with prior risk analyses and to identify emerging risks. Some risk management standards emphasise ongoing monitoring and periodic review as transversal elements of risk management (ISO, 2018; ISO/IEC, 2023), while others highlight the importance of comparing predicted and actual outcomes with regards to risk analysis more specifically (IEC, 2019, 6.4.3). In the context of frontier AI, both the NIST AI RMF Generative AI Profile (NIST, 2024) and the EU GPAI Code of Practice (2025) explicitly require post-deployment or post-market monitoring as part of risk analysis, with frontier AI posing several challenges in this respect. In practice, post-deployment monitoring for frontier AI systems draws on three main categories of information: (1) model integration and usage data, (2) application-level usage data, and (3) impact and incident data (Stein et al., 2024; Tanjaya & Pratt, 2025).

Model integration and usage data describe where and how models are deployed, disaggregated by sector, geography, and application type, but are currently limited and often reconstructed from voluntary self-reports, surveys, national accounting frameworks, and public registries (Highfill et al., 2025; Mora-López et al., 2025; Municipality of Amsterdam, n.d.; Tamkin et al., 2024). Some model developers, such as Anthropic and OpenAI, have released high-level anonymised usage statistics, (Anthropic, 2026; OpenAI, 2025c; Tamkin et al., 2024), but disclosure is overall limited (Wan et al., 2025). Application-usage data encompasses information about how users interact with applications deploying models in the real world, and could include AI application and AI agent activity logs or indices (MIT, 2025), user feedback channels, and coordinated flaw reporting by application developers (Cattell et al., 2025; Chan et al., 2024; NIST, 2024). These practices, however, are inconsistently applied, rarely standardised, and may raise unresolved privacy concerns (Stein & Dunlop, 2024; Tanjaya & Pratt, 2025). Coordinated mechanisms for reporting application flaws to centralised bodies could help regulators and civil-society organisations identify patterns of failure and target interventions (Gailmard et al., n.d.; Longpre & Appel, 2025; Richards et al., 2025). Impact and incident data capture real-world harms and broader social or economic effects, including reports of adverse incidents and sociotechnical field evaluations (Agarwal & Nene, 2024b). Voluntary repositories such as the AI Incident Database (AIID, n.d.) and OECD AI Incidents Monitor (OECD, n.d.) mirror practices in other safety-critical sectors (e.g., aviation and cybersecurity) but lack interoperability and mandatory participation (Agarwal & Nene, 2024b; McGregor, 2021; Stein et al., 2024; Turri & Dzombak, 2023). Sociotechnical field evaluations have generated meaningful insights into foundation model impacts on users (Tahaei et al., 2023; Vaccaro et al., 2024; Zhao et al., 2024) and subsequently informed model developers safety protocols (Hendrix, 2025). Yet, more robust analysis of model impacts and incidents will require sustained funding and interdisciplinary collaboration (Bengio et al., 2025).

In addition to reporting, post-deployment monitoring also includes model forensics: traceability techniques, such as embedding identifiable patterns into model outputs to make models uniquely traceable (Boenisch, 2021; Christ et al., 2024; Fernandez et al., 2023; X. Xu et al., 2024; N. Yu et al., 2021), and watermarking methods which help to either uniquely identify certain models or verify that content is generated from a specific model (Block et al., 2025; Gloaguen et al., 2025; L. Li et al., 2023). Metadata standards can also record contextual traces such as time, device, and location (Khan et al., 2018). Further research is needed to evaluate how to standardise and integrate traceability tools into post-deployment monitoring practices, balance traceability with privacy considerations, and how regulators and auditors might use model forensics to attribute responsibility for reported harms (Hilgert et al., 2025; Klasén et al., 2024).

> **Open Problems**
> 1. **How can external assessments be designed to provide sufficiently deep and meaningful access for rigorous evaluation, while protecting against misuse, intellectual property loss, and disclosure risks?**
>    *Who: Frontier AI developers' security and evaluation teams, third-party evaluators, AI audit and evaluations researchers, Standards bodies and regulators defining external assessment requirements*
>    **Type:** Lack of Consensus
>
> 2. **How to incentivise the institutionalisation of genuinely independent and diverse external assessment without reproducing capture, dependency, or concentration of evaluative power dynamics?**

> **Who:** *Third-party evaluators organisations, Regulators and public authorities designing oversight and funding mechanisms, Researchers in AI governance and audit design*
> **Type:** Shortcomings in implementation or application
>
> 3. **How can post-deployment monitoring for frontier AI systems systematically collect and link data on model integration, application usage, and real-world impacts to inform and review risk analysis results?**
>    *Who: Frontier AI developers, Downstream deployers, Risk management experts from other relevant sectors (e.g., cybersecurity, aviation), Intergovernmental organisations, Independent non-profit research organisations and affected stakeholders*
>    **Type:** Misalignment with or challenges to traditional risk management
>
> 4. **How can incentive structures be designed to encourage comprehensive post-deployment monitoring given that current practices remain largely voluntary, fragmented, and weakly standardised?**
>    *Who: Frontier AI developers, Downstream deployers, Regulators and intergovernmental organisations, Standards bodies*
>    **Type:** Shortcomings in implementation or application
>
> 5. **How can model forensics and traceability tools be integrated into post-deployment monitoring in ways that support accountability for harms while balancing privacy, security, and proprietary concerns?**
>    *Who: Frontier AI developers' security teams, AI forensics and security researchers, Regulators, Standards bodies*
>    **Type:** Lack of Consensus

## 3.3 Severity of Consequences and Likelihood

In safety risk management standards, risk analysis entails determining the level of risk, expressed as some combination of the severity of potential consequences (or harm) and the likelihood of those consequences (ISO/IEC, 2014). All the information collected internally and externally can be used, as appropriate, to arrive at an understanding of the relationships within and between different risks and to determine the level of risk. Below, we review open problems related to modelling risks and their interdependencies, as well as choosing the appropriate measures for risk.

**Modelling risks and interdependencies.** Modelling risks and interdependencies entails systematically analysing how identified risk sources, events, and system properties combine and propagate to produce real-world harms. In established high-risk industries, risk modelling is a central practice for safety risk management. For example, nuclear risk modelling combines Fault Tree Analysis (top-down deductive analysis tracing undesired events to root causes), with Event Tree Analysis (bottom-up mapping of potential outcomes following initiating events) (US NRC, 2016), while cybersecurity threat modelling scopes the scenario space by adopting an attacker's perspective to identify assets, trust boundaries, and plausible attack vectors (OWASP, 2025). In the context of frontier AI, risk modelling[5] is increasingly referenced in regulatory frameworks, such as the EU General-Purpose AI Code of Practice (2025), and is used to analyse how risks could materialise into concrete harms (Campos et al., 2025). Supported by techniques such as scenario analysis, this step connects technical system assessment with broader societal risk assessment by tracing causal chains linking model properties, user behaviour, and downstream impacts (Wisakanto et al., 2025).[6] Compared to mature practices in other safety-critical sectors, frontier AI risk modelling remains nascent, although several approaches are emerging.

Recent work adapts and extends methods from safety-relevant domains to account for frontier AI-specific characteristics. For example, Rodriguez et al. (2025) model how AI lowers attacker costs by identifying representative attack scenarios and bottlenecks in attack chains across threat landscapes, and quantifying how AI assistance reduces the costs of these bottlenecks. Wisakanto et al. (2025) adapt aerospace and nuclear safety techniques to AI by combining aspect-oriented hazard analysis, risk pathway modelling, bidirectional analysis from both capabilities and harms by combining techniques similar to event-tree and fault-tree analysis, and propagation operators that capture how risks may be amplified through accumulation, adversarial use, and sociotechnical diffusion. Building on this work, Murray et al. (2025b) propose a framework for risk modelling and quantification that selects representative scenarios, decomposes them into parameterised event sequences, establishes non-AI baselines, identifies benchmarks and indicators as proxies to estimate scenario parameters, and aggregates individual parameter estimates using statistical tools into high-level risk estimates, an approach applied to nine cybersecurity risk models (Barrett et al., 2025).

Despite this progress, several open problems remain. A central challenge is ensuring that risk models are sufficiently comprehensive, including capturing interdependencies across components, pathways, and sociotechnical dynamics (Shostack, 2014). Risk models must integrate heterogeneous evidence streams, such as usage and API data, incident reports, capability evaluations, and uplift studies measuring how AI systems enhance human ability to cause harm. Further research is needed on methods for systematically combining these inputs into unified models (Murray, Barrett, et al., 2025b). Additional challenges arise where historical data are sparse or absent, particularly for low-probability, high-severity risks In such cases, techniques such as expert elicitation (IEC, 2019, B.1) may be necessary, but their appropriate use and limitations for frontier AI remain underexplored.

**Quantifying risks.** Quantifying risks entails selecting appropriate metrics and scales to represent risks and their components in a form that supports comparison and decision-making. Risk management standards recognise that risk can be expressed using qualitative, semi-quantitative, or quantitative measures, depending on context and purpose (IEC, 2019). Common techniques for combining qualitative values include index methods and consequence–likelihood matrices (IEC, 2019, B.10.3), while a quantitative measure of risk can be produced from a probability distribution of consequences (e.g., VaR, CVar and S-curves). Risk management further emphasises that different risk values should be expressed on comparable scales, they need not be expressed through a single value and that the format of risk representation is appropriate for the risk at hand (IEC, 2019). Additionally, while quantitative scores may be easier to handle as they allow for easier comparisons and aggregation, they can also be misleading if used inappropriately (ISO, 2018), compressing uncertainty, and creating a false sense of precision, particularly when used for cross-domain comparisons or resource allocation (Aven & Reniers, 2013; Cox, 2008).

---

[5] In frontier AI risk management, mapping out consequences in a manner similar to scenario analysis is sometimes also referred to as threat modelling. It is important to note that this is very different from threat modelling as it has been originally understood in cybersecurity. In cybersecurity, threat modelling means identifying threats posed to the system and its possible consequences to the stakeholders of the system; whereas in the context of frontier AI risk management, threat modelling sometimes refers to identifying the threats posed by the AI system and its consequences on broader society in general.

[6] For example, a simplified cyber risk model might describe how moderately resourced cyber-attack groups target SMEs with ransomware by using AI to automatically harvest targets' emails, generate malware, and craft convincing phishing messages. By reducing both the expertise required and the time needed at each step, AI assistance can increase both the frequency and success rate of attacks, resulting in greater economic loss for SMEs. This illustrates how causal pathways can be traced from specific AI capabilities (e.g., code generation, text synthesis), through misuse scenarios, to concrete harms (e.g., financial losses, data breaches).

There is limited research explicitly referring to risk measures or estimates in frontier AI (Murray, Papadatos, et al., 2025; Wisakanto et al., 2025), with some work only mentioning measures for evaluation scores with little or no methodological explanation on how measures are derived (DSIT, 2024b; Solaiman et al., 2024; Weidinger et al., 2023). Specific examples of metrics include elicitation probabilities and capability scores (e.g., Ho et al., 2025) or the recently proposed 50%-task-completion-time-horizon metric for long tasks (Kwa et al., 2025). This metric choice is also important once probabilistic structures are applied because the resulting estimates are only meaningful if the metrics capture a relevant aspect of risk. Additionally, different frontier AI risks present distinct measurement challenges. Some risks, such as discrimination or human-rights harms, are difficult to capture through clear estimates (Yeung, 2025), even when using semi-quantitative or qualitative approaches. Others, such as loss of control, are hindered by the absence of relevant historical data (Chin, 2025). While documents like the EU GPAI Code of Practice (2025) provide high-level examples of frontier AI risk estimates,[7] specific guidance is lacking on how to present these in ways that remain interpretable and decision-useful while faithfully representing relevant aspects such as uncertainty and the plurality of possible harm pathways, rather than encouraging premature closure around simplified risk scores. As frontier AI risks may not admit a single dominant 'harmful scenario,' for one risk but rather a family of interacting scenarios (Barrett et al., 2025), it is unclear how these should be expressed in a useful and clear way through a single or multiple risk measures.

> **Open Problems**
> 1. **How can frontier AI risk models be made sufficiently comprehensive and successfully capture interdependencies, especially given the complexity yet limited historical data of some frontier AI risks?**
>    *Who: Frontier AI developers' safety and risk modelling teams, AI safety and risk modelling researchers, Domain experts in relevant fields (e.g. cybersecurity, biosecurity)*
>    **Type:** <u>Lack of Consensus</u>
>
> 2. **How can evidence from multiple and diverse data sources (e.g., evaluations, usage data, incident reports, and uplift studies) be systematically combined into coherent and decision-relevant risk models for frontier AI systems?**
>    **Who:** *Frontier AI developers' safety and risk modelling teams, Risk modelling researchers, Downstream deployers, Third-party evaluators*

---

[7] The EU GPAI Code of Practice asks that risk estimates are expressed as a risk score, risk matrix, probability distribution, or in other adequate formats, and may be quantitative, semi-quantitative, and/or qualitative. It explicitly cites examples such as a qualitative systemic risk score (e.g. 'moderate' or 'critical'); a qualitative systemic risk matrix (e.g. 'probability: unlikely' x 'impact: high'); and/or a quantitative systemic risk matrix (e.g. 'X-Y%' x 'X-Y EUR damage').

> **Type:** Misalignment with or challenges to traditional risk management
>
> 3. **How should frontier AI risk estimates be presented in an interpretable and decision-useful way while faithfully representing relevant aspects such as uncertainty, and the plurality of possible harm pathways?**
> *Who*: Frontier AI developers' safety and risk modelling teams, Risk modelling and metrology researchers, Domain experts in relevant fields (e.g. cybersecurity, biosecurity)
> **Type:** Misalignment with or challenges to traditional risk management

# 4. Risk Evaluation

Risk evaluation involves the process of comparing the results of risk analysis (Section 3) with the criteria set during the planning stage (Section 1.3) in order to determine whether the risk is acceptable (ISO, 2022a, 3.3.25). Safety risk management makes explicit reference to risk acceptance, by linking risk evaluation to the goal of determining whether acceptable risk has been exceeded (ISO/IEC, 2014). If risk is deemed unacceptable, risk mitigation ought to be pursued (Section 5) and risk needs to be re-evaluated after another round of risk assessment, until risk is acceptable. If risk is deemed acceptable, it is possible to proceed with product release. Below, we review open problems related to determining risk acceptance and, when risk is deemed acceptable, making deployment decisions.

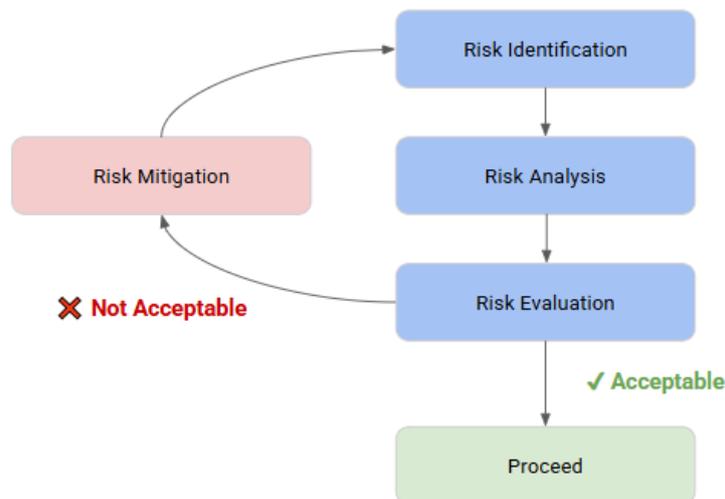

*Figure 2. Iterative process of risk evaluation*

## 4.1 Determining Risk Acceptance

The data obtained through risk analysis (Section 3) can be used to inform decisions about whether the risk should be accepted or whether it requires mitigation, and if so, any priorities for mitigation. Below, we review relevant open problems.

**Applying risk acceptance criteria.** Risk criteria are applied to determine the significance of the risk relative to the criteria set by the organisation beforehand (ISO, 2022a, 3.3.6) and, based on those criteria, decide whether the risk should be accepted or mitigated (ISO/IEC, 2014). Traditional safety risk management techniques, such as ALARP (Section 1.3), provide a way to distinguish between intolerable risk that cannot be justified, risk that should be reduced where reasonably practicable, and risk considered sufficiently low to be accepted without further treatment (IEC, 2019). In frontier AI, guidance (Raman et al., 2025) and regulatory documents such as the EU GPAI Code of Practice (2025) feature the application of risk acceptance criteria, or 'thresholds', requiring also the incorporation of a safety margin. The EU GPAI Code of Practice specifies that such a safety margin should account for potential changes, uncertainties and limitations related to risk sources (e.g., post-assessment capability improvements), the assessment process itself (e.g., under-elicitation in evaluations), and the effectiveness of mitigations (e.g., circumvention or deactivation) (2025). As mentioned in Section 1.3, the best practice among frontier AI companies is to currently rely on, and thus apply, capability thresholds to determine risk acceptance. While, as mentioned in Section 2.1, companies tend to focus on a similar set of risk domains for analysis (e.g., CBRN), the application of capability thresholds varies across companies. Some developers use thresholds as 'tripwires' leading to outcomes such as 'do not release' or 'stop development' only mentioning mitigations and measures at a high-level (Meta, 2025), while others apply them as mitigation-and-decision ladders, triggering specific safety and security standards (Anthropic, 2025; OpenAI, 2025d). Additionally, they often fail to incorporate safety margins as required in frontier AI regulatory documents, thus not accounting for potential uncertainties in assessment, effectiveness of mitigations or unexpected changes in risk sources. This uneven application makes it difficult for regulators, policymakers, safety or security professionals, end-users and downstream developers to clearly understand the risks at hand and the choices made by model developers (e.g., choices around mitigation or deployment). Consistency in the application of risk thresholds and risk acceptance criteria, as is the norm in other critical industries such as aviation, could help alleviate this (Campos et al., 2025).[8]

**Determining overall risk acceptance.** Determining overall risk acceptance entails aggregating multiple individual risks to form a judgment about whether the system's total risk profile is acceptable. Traditional risk management emphasises risk aggregation as a means of combining individual risks to inform holistic acceptance decisions (ISO, 2022a, 3.3.30). Aggregation methods range from relatively simple approaches, such as weighted sum of all risks with the possibility of context-specific weighting for the most relevant hazards (Schmitz et al., 2025), to approaches that explicitly account for interdependencies between risks (IEC, 2019). Established practices also recognise trade-offs across risks, for example through concepts such as 'globally at least equivalent,' (GALE) whereby a risk with adverse consequences may be deemed acceptable if equivalent or greater risk reductions have been achieved elsewhere (IEC, 2019). In frontier AI, however, overall risk acceptance remains underdeveloped. While existing developers' safety frameworks tend to focus on acceptance decisions at the level of individual risks, the aggregation of multiple AI risks (and the criteria by which aggregate acceptance should be carried out) are often absent or left implicit. Frontier AI presents distinct challenges for risk aggregation, and there is limited literature examining whether traditional aggregation methods meaningfully transfer to frontier AI contexts (Schmitz et al., 2025). Traditional factors such as interdependencies between risks and information loss during aggregation (David, 2009), and AI-specific challenges such as diverse application contexts and differing degrees to which various risks can be quantified, complicate aggregation efforts (Schmitz et al., 2025). Frontier AI risks span heterogeneous domains, including economic harms (e.g., AI-enabled cyber attacks), physical harms (e.g., CBRN uplift), diffuse societal harms (e.g., manipulation or erosion of trust), and rights-based harms (e.g., discrimination or privacy violations), which may seem incommensurable in the face of aggregation. This complexity leaves it especially unclear how single risk evaluations should inform overall risk acceptance, specifically whether the presence of a single unacceptable risk should render the overall model risk unacceptably high.

---

[8] In the aviation industry, it is the Federal Aviation Administration, rather than individual companies, that sets the acceptable frequency of catastrophic accidents.

> **Open Problems**
> 1. **How can risk acceptance criteria be made to be applied more consistently across frontier AI model developers so that risk acceptance and safety priorities are comparable and interpretable to external stakeholders?**
>    *Who:* *Frontier AI developers' safety and policy teams, Standards bodies, Regulators and intergovernmental bodies, Third-party evaluators*
>    **Type:** Shortcomings in implementation or application
>
> 2. **By which criteria should aggregate risk acceptance be carried out given differences in measurability and interdependencies between different risks?**
>    **Who:** *Frontier AI developers' safety and policy teams, Risk modelling and socio-technical researchers, Domain experts from specific risk-relevant fields, Standards bodies*
>    **Type:** Misalignment with or challenges to traditional risk management

## 4.2 Deployment Decisions

The results of risk evaluation inform product-release decisions, where risk has been deemed acceptable. In the case of frontier AI, this refers to deployment decisions; the act of releasing AI systems as standalone models, real-world applications or services. Below, we review relevant open problems.

**Deployment decisions protocols.** Deployment decision protocols determine whether, how, and under what conditions an AI system is released or made available for use (Bengio et al., 2026). Risk management standards emphasise that such decisions should follow a continuous and iterative process of risk evaluation and analysis, including reassessment after mitigations are implemented (ISO, 2018; ISO/IEC, 2014, 2023). In frontier AI, this approach is reflected in regulatory and governance guidance (EU Commission, 2024, 2025), outputs from AI Safety Summits (G7, 2023), and independent research (Barrett et al., 2025; Kaminski, 2023). This means that risk should be re-assessed after risk mitigations have been implemented and, in cases where risk is still deemed unacceptable, developers should implement additional mitigations until risk levels are acceptable, halt deployment, or recall deployed models from the market (EU Commission, 2025; J. O'Brien et al., 2023). Frontier AI raises specific challenges for deployment decisions due to uncertainty about downstream impacts and the diversity of possible release strategies. Transparency around deployment and release decisions is therefore particularly important (Bommasani et al., 2023), as different risk levels may warrant different strategies (Anderljung et al., 2023), such as gradual release, gated to public access, or hosted access, among others (P. Liang et al., 2022; Seger et al., 2023; Solaiman, 2023). While transparency around external releases has increased in recent years (Wan et al., 2025), far less is known about the factors and protocols that shape internal deployment decisions (Stix et al., 2025), where a developer develops a model or system and makes it available exclusively for internal access or use, with some researchers suggesting that internal governance mechanisms for such decisions may be largely absent (Stix et al., 2025). Because the most cutting-edge AI systems are often deployed internally first, and given questions around potential regulatory gaps in certain jurisdictions (Pistillo, 2026), these deployments can pose significant under-scrutinised risks. Researchers suggest to adapt lessons from other safety-critical industries (e.g., biological agents and toxins, R&D in nuclear reactors, novel medical devices, etc.) to establish internal use policies, oversight frameworks for internal deployment decisions, and appropriately targeted transparency mechanisms to minimise potential risks coming from such deployments (Stix et al., 2025).

**Justifying deployment decisions**. Justifying deployment decisions entails documenting and validating the reasoning for proceeding with deployment based on the outcomes of risk evaluation. Risk management standards require that risk assessment results be recorded, communicated, and reviewed at appropriate organisational levels (ISO, 2018). In safety-critical domains, this justification often includes demonstrating that risks cannot be reasonably reduced further prior to deployment (ISO/IEC, 2014). Overall, this step is useful to justify decision-making, as well as signaling compliance to regulators (Liberati et al., 2024). In frontier AI, some regulatory frameworks (EU Commission, 2025) specifically require that companies provide their reasons for proceeding with deployment in model reports. In the last few years, there has been an increasing focus on showing that a frontier AI system is sufficiently safe to justify its deployment, and proposals for suitable approaches, such safety cases, have emerged. Common across numerous industries (Leveson, 2011a; Maguire, 2017; Sujan et al., 2016), safety cases provide a structured argument, supported by a body of evidence (e.g., results of risk assessment) that an AI system is safe to deploy in a specific setting.

Safety cases are provided by frontier AI developers, and could serve different objectives depending on their scope. While a few developers have begun using safety cases to address only partial or specific risks (e.g., Anthropic's 'affirmative cases' in their Responsible Scaling Policy (2025) or Google DeepMind's use of safety cases in their Frontier Safety Framework (2025)), some propose using safety cases more broadly to offer a rationale for why the probability of an AI system causing a catastrophe is below an acceptability threshold during a deployment window (Balesni et al., 2024; Clymer et al., 2024; Irving, 2024). Stakeholders across government (M. Buhl et al., 2025), industry (Anthropic, 2025; Google, 2025) and the research community (Y. Bengio et al., 2024; M. D. Buhl et al., 2024) have recommended using safety cases as a key input to deployment decisions. However, approaches to constructing safety cases for frontier AI remain nascent (Buhl et al., 2025; Korbak et al., 2025) and guidance on how results from risk analyses should be integrated into safety cases is still underdeveloped. On a more fundamental level, it is an open question whether safety cases are the right approach for a fast-changing technology such as frontier AI. Designed for mature technical systems (Bishop & Bloomfield, 1998), the value of safety cases diminishes when risks are numerous, ill-defined, or hard to model (Leveson, 2011b). Since safety cases cannot rule out unknown risks and that, as presented in Section 2, current approaches in frontier AI struggle to capture AI hazards comprehensively and in detail, relying on them may create a false sense of safety, especially for unpredictable but high-impact failures.

---

**Open Problems**
1. **How should appropriate transparency and rigorous internal governance mechanisms be ensured around internal deployment decisions for frontier AI?**
   *Who:* Frontier AI developers' safety and policy teams, Third-party evaluators, Regulators
   **Type:** Shortcomings in implementation or application

> 2. **Under what conditions are safety cases an appropriate mechanism for justifying deployment decisions in frontier AI, especially in the face of uncertain and hard-to-model risks?**
> **Who:** *Frontier AI developers' safety and evaluation teams, Safety engineering experts, AI safety and evaluation researchers, Standards bodies, regulators*
> **Type:** Misalignment with or challenges to traditional risk management

# 5. Risk Mitigation

Risk mitigation refers to risk treatments that deal with negative consequences, also referred to as 'risk reduction' (ISO, 2022a, 3.3.32). Risk mitigation aims to bring the level of risk, as evaluated in Section 4, to an acceptable level (ISO/IEC, 2014). Risk management in safety-critical sectors, such as nuclear (IAEA, n.d.), organises risk mitigations according to a hierarchy (ISO/IEC, 2014). The hierarchy provides essential guidance by emphasising the relative effectiveness of mitigations, starting from inherently safe design, to guards and protective devices down to information for use (ISO/IEC, 2014) (See *Figure 3*). The assumption behind it is that protective measures inherent to the characteristics of the product or system are likely to remain effective, whereas even well-designed guards and protective devices can fail or be violated, and information for use might not be followed (ISO/IEC, 2014). Without this framing, the treatment of risk management from a safety perspective can appear incomplete.

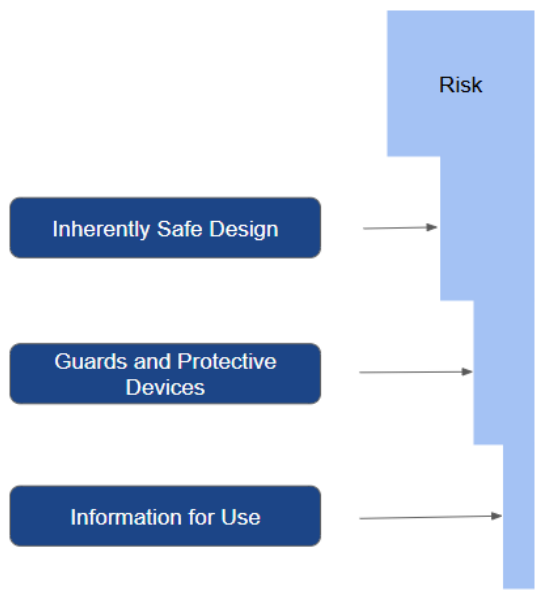

*Figure 3. Three-step method (or hierarchy) of risk reduction (Adapted from* (ISO/IEC, 2014)*, design phase)*

In order to reasonably align with the principle of risk reduction above, we attempt to present frontier AI safety mitigations according to the level at which they reduce risk, and thus their relative effectiveness, ordering the following sections from data-, model-, system- up to ecosystem-level mitigations. Given how specific to AI mitigations are, the reference to existing standards is unavoidably reduced in the following paragraphs. Additionally, we constrain the scope to mitigations related to model and system safety, and consider security concerns out of scope for this iteration. Below, we review open problems.

## 5.1 Data-level mitigations

In risk management, measures that act on the inherent characteristics of the product are the first and most important step in the risk mitigation process. Below, we review data-level mitigations and their open problems accordingly.

**Data-level mitigations.** Data-level mitigations aim to reduce risk by constraining the emergence of hazardous capabilities intervening on the data that the model is trained on. From a risk management perspective, such mitigations are appealing because they intervene early in the model lifecycle, recalling the idea of inherently safe design (ISO/IEC, 2014). However, their value depends on whether they can deliver reliable and demonstrable risk reduction in practice. While frontier AI systems benefit from broad knowledge and capabilities, certain types of knowledge pose safety risks, such as language models knowing how to assist in cyber, bio, or chemical attacks; or image/video models 'knowing' how to create nonconsensual intimate deepfakes. An intuitive way to avoid having models learn unwanted capabilities is to filter the data they are trained on, particularly during the pretraining process when models develop core representations of knowledge. However, filtering training data is deceptively difficult (Paullada et al., 2021) due to costs (Ngo et al., 2021), filtering errors (Ziegler et al., 2022), degradation of dataset quality (Welbl et al., 2021), the massively multilingual nature of internet text (Kreutzer et al., 2022), cultural biases in content moderation (Dodge et al., 2021; Stranisci & Hardmeier, 2025; Welbl et al., 2021; A. Xu et al., 2021), and the inherently contextual nature of harmful behaviour. Frontier AI introduces challenges related to the limits of controllability at the data level, with emerging evidence suggesting that models may be able to robustly lack knowledge in complex domains such as science and engineering (B. W. Lee et al., 2025; K. O'Brien et al., 2026) but not simpler tendencies such as toxicity (K. Li et al., 2025; Maini et al., 2025). This casts questions on the ability to control for safety at the source when it comes to frontier AI and thus, also on the ability to select mitigations according to their expected effectiveness, thus respecting a classic safety risk-management logic, a priori. Ultimately, it is an ongoing challenge to develop effective methods for data curation, characterise the relationship between training data contents and emergent capabilities, and make models that more robustly lack harmful abilities (Barez et al., 2025; Casper, O'Brien, et al., 2025).

> **Open Problems**
> 1. How can data-level mitigations for frontier AI be made into effective risk controls, given the uneven controllability of different capabilities and behaviours through such interventions?
> *Who:* Frontier AI developers' pre-training and data curation teams, AI safety and mitigations researchers, and third-party evaluators, safety engineering experts
> **Type:** <u>Misalignment with or challenges to traditional risk management</u>

## 5.2 Model-level mitigations

There are also a host of technical engineering controls that act at the model level, and which aim to reduce assessed risk by shaping and constraining model behaviour. Below, we review them and related open problems.

**Model behaviour mitigations.** Beyond intervening on the data, and thus on the knowledge representations that the model is learning, another set of mitigations can constrain model behaviour after such representations have been learned. This may include a host of different approaches, from fine-tuning to machine unlearning. To public knowledge, the state-of-the-art for fine-tuning frontier AI systems in recent years have been methods that rely on feedback or demonstrations from humans or AI systems (e.g., Bai et al., 2022; Kaufmann et al., 2025; C. Zhou et al., 2023). However, the effectiveness of these methods is limited by the quality of feedback and demonstrations provided to the AI system (e.g., Casper et al., 2023; Lambert & Calandra, 2024; Lindström et al., 2024). Not only are humans prone to simple mistakes and disagreement (Glickman & Sharot, 2025; M. Wu & Aji, 2023), but optimising for human approval can systematically, and often subtly, cause systems to learn harmful behaviours. For example, frontier language models are prone to be sycophantic to users, pandering to their preferences at the expense of objectivity and truth (Malmqvist, 2024; OpenAI, 2025b; Perez, Ringer, et al., 2022; Sharma, Tong, Korbak, et al., 2025). This can be understood as a form of `reward hacking:' the phenomenon in which AI systems can learn perverse behaviours by gaming imperfect reward signals (Baker et al., 2025; Skalse et al., 2022). It is also challenging, even for human experts, to oversee LLM's performance on very difficult and complex tasks such as spotting vulnerabilities in a complex codebase or errors in an advanced proof (Kim et al., 2024). In response to these challenges, researchers are working on methods for human-AI teams to help evaluate complex behaviours (Du et al., 2023; Kenton et al., 2024; A. Khan et al., 2024; McAleese et al., 2024; Michael et al., 2023; Wen et al., 2026).

Beyond increasing complexity of behaviour, frontier AI introduces additional challenges due to emergent and longitudinal harms. Harm does not always result from single uses of AI systems, but often from the sum total of a system's behaviour across contexts and its effects on users. For example, after a 16-year-old committed suicide in April 2025, conversations with ChatGPT revealed the model provided harmful advice, instructions, and 1,275 mentions of suicide over the course of months in many separate chats (Tabachnik, 2025). Meanwhile, emerging research is beginning to identify harmful emergent effects of AI in education (e.g., Tamimi et al., 2024), mental health (e.g., Caridad, 2025), and user judgment (e.g., Krügel et al., 2023). From a risk management perspective, this complicates risk mitigation. Many model behaviour controls are evaluated on short-term interactions, while the most severe risks materialise gradually across sectors, and are increasingly difficult to control for and evaluate.

Aside from techniques focused on aligning model behaviours with human interests, another technique for making AI systems safer is to use 'machine unlearning' algorithms to remove harmful capabilities (Barez et al., 2025; S. Liu et al., 2024). Existing unlearning techniques (e.g., Sheshadri et al., 2025; Zou et al., 2024) are effective at suppressing harmful capabilities in most situations, but adversarial users can easily prompt, fine-tune or otherwise attack the model to resurface these capabilities (Che et al., 2025; Cooper et al., 2025; Deeb & Roger, 2025; Hu et al., 2025; Huang et al., 2024; Lo et al., 2024; Łucki et al., 2025; Qi et al., 2024; B. Wei et al., 2025). A significant challenge is designing unlearning algorithms that result in more robust knowledge removal, possibly with algorithmic innovations or scaling unlearning interventions (Casper, O'Brien, et al., 2025). For risk management, the key open problem is durability: reversible or fragile mitigations risk creating unjustified confidence, leading to deployment decisions that are not supported by sustained risk reduction.

**Model-robustness.** Model-robustness addresses the risk that safety failures arise under distributional shift or adversarial conditions. Despite their intelligence, even state-of-the-art AI systems are prone to exhibiting both subtle and egregious failures. One problem is building models that generalise appropriately and predictably outside of their training data distribution. For example, modern language models tend to be less safe (Shen et al., 2024; Song et al., 2024; W. Wang et al., 2024; Yong et al., 2024) and performant (Kshetri, 2024; Salammagari & Srivastava, 2024) in low-resource languages. Meanwhile, modern LLMs are vulnerable to especially egregious safety failures in the face of adversarial attacks, such as 'jailbreaks' which trick these models to comply with arbitrary harmful requests (Chowdhury et al., 2024; Jiang et al., 2024; Jin et al., 2025; A. Wei et al., 2023; Yi et al., 2024). Even cutting-edge systems are persistently vulnerable to attacks. For example, a recent public competition to crowdsource attacks compiled over 60,000 successful attacks against recent production models from Amazon, Anthropic, Cohere, Meta, Mistral, OpenAI, and xAI (Zou et al., 2025). The principal challenge for improving adversarial robustness is to drive attack success rates down. Doing so will benefit from innovations in red-teaming, increasing the scale of adversarial training (Howe et al., 2025; Lee et al., 2025), and developing new algorithms for adversarial robustness (Casper, O'Brien, et al., 2025; Casper, Schulze, et al., 2025; Dékány et al., 2025; Fu & Barez, 2025; Sheshadri et al., 2025; Zou et al., 2024), on which more work is needed. Red teaming, for example, while useful through the open-ended use of adversarial attacks or interpretability techniques (e.g., Marks et al., 2025) is skill-dependent and frequently fails to be consistently rigorous in practice (Feffer et al., 2024). Red-teaming methods regularly empirically fail to identify failures before deployment and thus, improving model robustness will require improvements in the adversarial capability elicitation toolkit (e.g., Che et al., 2025; Lüdke et al., 2025). This creates a decision-making challenge under incomplete evidence, where organisations must decide how to act when robustness assessments provide only partial or negative assurance. Improving robustness therefore requires not only technical advances, but clearer guidance on how robustness evidence should inform risk acceptance and the selection of complementary mitigations where needed.

> **Open Problems**
> 1. **How can model behaviour controls be used as reliable risk mitigations when growing system behavioural complexity and emergent cross-sectoral harms undermine our ability to evaluate their effectiveness?**
>    *Who:* *Frontier AI developers' safety and mitigation teams, AI safety mitigations and alignment researchers, safety engineering and risk management experts*
>    **Type:** Misalignment with or challenges to traditional risk management
>
> 2. **How can model-level mitigations, such as machine unlearning, ensure durability and be relied upon for sustained risk reduction?**
>    **Who:** *Frontier AI developers' research and safety mitigation teams, Researchers in AI safety mitigations, safety engineering experts*
>    **Type:** Lack of Consensus
>
> 3. **How can partial or negative model robustness evidence be accounted for, rather than mislead, risk acceptance and the potential selection of complementary mitigations?**
>    **Who:** *Frontier AI developers' safety mitigations and robustness teams, Researchers in AI model robustness and safety mitigations, safety engineering and risk management experts*
>    **Type:** Misalignment with or challenges to traditional risk management

## 5.3 System-level mitigations

Frontier AI risk management increasingly relies on system safety architectures: a set of external mechanisms that monitor, steer, and constrain model behaviour without requiring the computationally expensive retraining of the underlying model. AI system safety treats the model as a component within a larger system where safety is enforced through system monitoring, inference-time control mechanisms, and system's guardrails. Below, we review related open problems.

**System monitoring.** Monitoring serves as the primary tool for identifying unsafe or otherwise unwanted behaviours. In risk management terms, monitoring primarily supports detection and escalation rather than direct risk reduction. Its effectiveness therefore depends on whether monitoring signals reliably trigger appropriate and timely interventions, such as human review, access restrictions, or system rollback. Monitoring techniques can vary by the object of oversight – focusing on hardware activity (Aarne et al., 2024; O'Gara et al., 2025), users (e.g., Brown et al., 2025; Yueh-Han et al., 2025), inputs/outputs (e.g., Greenblatt et al., 2024; McKenzie et al., 2026; Sharma et al., 2025), model uncertainty (e.g., Farquhar et al., 2024; Lamb et al., 2026), internal cognition (e.g., Goldowsky-Dill et al., 2025; Kirch et al., 2025; Kramár et al., 2026), and reasoning (e.g., Baker et al., 2025; Korbak et al., 2025) – or by the goal of oversight – logging information, flagging risky content, filtering harmful content, triggering failsafes, spotting deception, etc. With regards to frontier AI, a primary challenge in this domain is the creation of robust classification systems capable of identifying safety risks in real-time. One way to achieve this is through the deployment of specialised, instruction-tuned safety models that evaluate human-AI conversations against multi-class hazard taxonomies or are designed to detect toxic prompts and adversarial jailbreak attempts that might otherwise bypass general-purpose filters (e.g., AlDahoul et al., 2024; Han et al., 2024; Inan et al., 2023; Lee et al., 2025; Liu et al., 2025; Sharma, Tong, Mu, et al., 2025; Yuan et al., 2024; Zeng et al., 2024). However, as models and users co-evolve, it is increasingly difficult to ensure that monitoring signals remain up to date and do not degrade over time, triggering inappropriate escalation responses and wrong risk mitigation strategies.

**Control mechanisms.** Inference-time control mechanisms allow for real-time modification of model behaviour, enabling post-deployment alignment that is both lightweight and adaptive. From a risk management perspective, these mechanisms function as operational controls that seek to reduce the possibility of harm by constraining behaviour at the point of use. Some approaches include activation engineering, which involves manipulating the model's internal representations during the forward pass to guide outputs away from harmful behaviours or biases (Bayat et al., 2025; Ghosh et al., 2025; Postmus & Abreu, 2025; Turner et al., 2024). To achieve more precise and interpretable control, techniques that influence a model's exploratory behaviour and uncertainty (e.g., Rahn et al., 2024), modulate token-level probability distributions at the decoding stage (e.g., Chakraborty et al., 2025; Pynadath & Zhang, 2025; Wang et al., 2025), and model the generation process as a consensus-driven interaction between generators and verifiers (e.g., Welleck et al., 2024) have been developed. Regarding frontier AI, maintaining control over a model's behaviour using steering and control techniques is often difficult. Empirical studies have highlighted persistent failures in coverage and the emergence of unintended side effects when attempting to steer large-scale models (e.g., Bentley, 2025; Miehling et al., 2025). To address these shortfalls, frameworks that use episodic memory or flexible backtracking mechanisms to dynamically determine the necessity and strength of intervention throughout the generation process have been proposed (e.g., Cheng et al., 2025; Do et al., 2025; Wu et al., 2025). Still, as a risk mitigation measure, such mechanisms remain brittle and have yet to provide meaningful safety assurances that could support their choice for risk reduction.

**System's guardrails.** Guardrails act as the integrated system logic that triggers interventions when violation of safety objectives are imminent, functioning as a bidirectional filter between the user and the model. In high-stakes applications, guardrails can enforce rigid structural requirements, using neuro-symbolic frameworks constraints to ensure outputs adhere to formal logical rules (e.g., Zhang et al., 2024). Modern deployment scaffolding integrates these various components into modular API gateways that orchestrate multi-agent validation and enforce compliance with security policies and other external standards (e.g., Shvetsova et al., 2025). Another approach involves the use of defensive mechanisms designed to detect and recover from alignment drifting caused by poisoning attacks, prompt injections, or malicious fine-tuning (e.g., Liao et al., 2025; Yan et al., 2024). Such protective measures can be combined with human-centric safety filters that reason about the feedback loop between AI outputs and human behaviour to ensure long-term robustness (e.g., Bajcsy & Fisac, 2024). Ultimately, each type of AI system guardrail aims to embed constitutional safety principles into evolving deployment scenarios. Frontier AI introduces particular challenges for guardrails because deployment contexts, usage patterns, and model capabilities evolve rapidly. As systems scale and become more adaptive, maintaining reliable alignment between guardrail logic and real-world risk becomes increasingly difficult. Guardrails must operate under uncertainty, adversarial pressure, distribution and domain shifts, and changing objectives, raising the risk of both under-enforcement (allowing harmful behaviour) and over-enforcement (unduly constraining legitimate use).

> **Open Problems**
> 1. **How to ensure that system monitoring remains accurate and does not trigger inappropriate escalation responses as users and models co-evolve?**
>    *Who: Frontier AI developers' safety and mitigation teams, human-AI interaction and safety mitigation researchers, safety engineering experts*
>    **Type:** Misalignment with or challenges to traditional risk management
>
> 2. **How can AI system behaviour be effectively controlled or steered towards safety to ensure meaningful risk reduction?**
>    *Who: Frontier AI developers' research and alignment teams, Researchers in AI alignment and AI safety mitigations*
>    **Type:** Lack of Consensus
>
> 3. **How to maintain reliable alignment between guardrail logic and real-world risks as deployment contexts, usage patterns and capabilities evolve rapidly?**
>    *Who: Frontier AI developers' safety and deployment teams, AI users (.e.g, downstream deployers), safety engineering and risk management experts*
>    **Type:** Misalignment with or challenges to traditional risk management

## 5.4 Ecosystem-level mitigations

In frontier AI, ecosystem-level mitigations may also be understood more broadly as developers providing information, tools, and capabilities that enable other actors (e.g., governments, organisations, and civil society) to implement effective defenses against AI-enabled safety risks. Below, we review some of the open problems.

**Information-sharing and documentation.** In safety risk management, documentation does not reduce risk directly (ISO/IEC, 2014), but it plays a critical role in ensuring that relevant information about deployed systems, testing results and limitations, to cite a few, are sufficiently understood and shared across the ecosystem to prevent or minimise safety risks (FMF, 2025a). More specifically, in the context of frontier AI, documentation practices such as model cards (Mitchell et al., 2019), datasheets for datasets (Gebru et al., 2021), suppliers' declarations of conformity for AI services (Arnold et al., 2019), and now increasingly agent cards (OpenAI, 2025a), aim to surface information about training data, intended use, performance boundaries, and known failure modes to support safer development, (re-)use and accountability in development and deployment decisions. Despite their widespread adoption, existing documentation practices exhibit inconsistency in both content and quality of documentation, limiting their effectiveness as risk-mitigation tools (Richards et al., 2020; Staufer et al., 2025). Studies show uneven adherence to model and dataset documentation standards and significant variation in what information is disclosed (Liang et al., 2024; Staufer et al., 2025), with sections addressing environmental impact, limitations, and evaluation showing the lowest filled-out rates (Liang et al., 2024). From a risk-management perspective, this undermines the ability of downstream actors (such as safety teams, auditors, deployers, and regulators) to assess whether existing practices are appropriate, sufficient, or being applied under the assumptions for which they were designed. Moreover, most documentation frameworks remain poorly adapted to domain-specific requirements. For example, Datasheets for Datasets fall short in meeting domain specific requirements such as around medical data that are required for data documentation and screening prior to AI applications (Marandi et al., 2025). As a result, a key open problem at the risk mitigation stage is how to design documentation practices that reliably contribute to meaningful reduction of risks, rather than merely increasing transparency. This includes determining what information is necessary and sufficient to support mitigation decisions, and how to standardise documentation in ways that remain sensitive to sector-specific requirements (e.g., medical or safety-critical domains).

**Serious Incident Reporting.** In safety risk management, incident reporting does not prevent hazards ex ante, but it plays a critical role in limiting further harm, identifying systemic weaknesses in existing controls, and informing corrective actions that reduce future risk. In the context of AI, incident reporting encompasses formal processes for documenting and sharing safety incidents, security breaches, near-misses, and relevant threat intelligence with appropriate stakeholders to enable coordinated responses and systemic improvements (Saeri et al., 2025). While several public databases track AI incidents (including the AI Incident Database (AIID, n.d.), AIAAIC Repository (AIAAIC, n.d.), and MIT AI Incident Tracker (MIT, n.d.), to cite a few), formal reporting obligations remain uneven across jurisdictions, with mandatory reporting currently concentrated in the EU under the AI Act and its GPAI Code of Practice (EU Commission, 2024, 2025). Despite growing institutional attention, current incident reporting practices face structural limitations that constrain their effectiveness as risk-mitigation tools for frontier AI. Key challenges include the absence of shared definitions for what constitutes a reportable incident, and difficulties in capturing harms that unfold over time, recur across contexts, or manifest primarily at the societal level rather than as discrete events (Hoffmann & Frase, 2023; Paeth et al., 2024a). Inconsistent database structures and incompatible data fields further limit the ability to aggregate incidents, identify patterns, or link reported harms to specific model features, deployment decisions, or mitigation failures (Agarwal & Nene, 2024a; Dixon & Frase, 2024; OECD, 2025). These limitations are compounded by the largely voluntary nature of reporting outside the EU, which results in under-representation of developer-side failures (Li et al., 2025), coverage bias toward high-profile misuse cases and limited visibility into incidents occurring in less-resourced contexts (Agarwal & Nene, 2024a; Paeth et al., 2024b). Empirical evidence shows that only a small fraction of reported AI incidents trigger observable responses or corrective actions, highlighting a persistent gap between reporting and mitigation (Richards et al., 2025b). Addressing this gap requires clearer links between incident reports and concrete risk-treatment measures. Possible ways forward include requiring reports to include information on the status of damage mitigation (Prud'homme et al., 2023), creating emergency protocols for rollbacks (Uuk et al., 2024), or the establishment of escalation channels through which implementers, vendors, and regulators can be informed about critical vulnerabilities (Gipiškis et al., 2024b). However, with the increasing volume of reports, regulators are also facing potential regulatory overload (Cebrian et al., 2025).

> **Open Problems**
> 1. **How to design documentation practices that reliably contribute to meaningful and observable reduction of risks, rather than merely increasing transparency?**
>    *Who:* *Frontier AI developers' policy and documentation teams, regulators and intergovernmental bodies, standards bodies*
>    **Type:** Shortcomings in implementation or application
>
> 2. **How can serious incident reporting frameworks be designed so that reported incidents are consistently defined and meaningfully linked to concrete corrective actions, without overwhelming regulators or discouraging reporting?**
>    **Who:** *Frontier AI developers' safety and policy teams, regulators and intergovernmental bodies, standards bodies, incidents database operators, security experts from other relevant fields (e.g., cybersecurity)*
>    **Type:** Shortcomings in implementation or application

# 6. Conclusion

This paper has argued that existing risk management standards and AI safety practices are alone insufficiently equipped to address the distinctive challenges posed by frontier AI. While a growing number of initiatives seek to fill this gap, the absence of a shared, problem-oriented agenda risks fragmentation, duplication, and misalignment with established risk management principles. In response, we have systematically surfaced open problems across the core stages of the risk management process, focusing on organisational mechanisms, yet prioritising relevant safety aspects for Frontier AI. By classifying these problems and clarifying where consensus, alignment, or implementation remains lacking, this work aims to support more coordinated and effective progress. The resulting agenda-setting reference document and living repository are intended to help stakeholders prioritise efforts, foster convergence, and lay the groundwork for more robust and credible frontier AI risk management practices.

Preyssl, C. (1995). Safety risk assessment and management—The ESA approach. *Reliability Engineering & System Safety, Space System Applications of Risk Assessment*, *49*(3), 303–309. https://doi.org/10.1016/0951-8320(95)00047-6

Prud'homme, B., Régis, C., & Farnadi, G. (2023). *Missing Links in AI Governance*. UNESCO. https://unesdoc.unesco.org/ark:/48223/pf0000384787

Pynadath, P., & Zhang, R. (2025). *Controlled LLM Decoding via Discrete Auto-regressive Biasing* (arXiv:2502.03685). arXiv. https://doi.org/10.48550/arXiv.2502.03685

Qi, X., Wei, B., Carlini, N., Huang, Y., Xie, T., He, L., Jagielski, M., Nasr, M., Mittal, P., & Henderson, P. (2024). *On Evaluating the Durability of Safeguards for Open-Weight LLMs* (arXiv:2412.07097). arXiv. https://doi.org/10.48550/arXiv.2412.07097

Rahn, N., D'Oro, P., & Bellemare, M. G. (2024). *Controlling Large Language Model Agents with Entropic Activation Steering* (arXiv:2406.00244). arXiv. https://doi.org/10.48550/arXiv.2406.00244

Raji, I. D., Xu, P., Honigsberg, C., & Ho, D. (2022). Outsider Oversight: Designing a Third Party Audit Ecosystem for AI Governance. *Proceedings of the 2022 AAAI/ACM Conference on AI, Ethics, and Society, AIES '22*, 557–571. https://doi.org/10.1145/3514094.3534181

Raman, D., Madkour, N., Murphy, E. R., Jackson, K., & Newman, J. (2025). *Intolerable Risk Threshold Recommendations for Artificial Intelligence* (arXiv:2503.05812). arXiv. https://doi.org/10.48550/arXiv.2503.05812

Renieris, E. M., Kiron, D., & Mills, S. (2024, April 23). *AI-Related Risks Test the Limits of Organizational Risk Management*. MIT Sloan Management Review. https://sloanreview.mit.edu/article/ai-related-risks-test-the-limits-of-organizational-risk-management/

Reuel, A., Bucknall, B., Casper, S., Fist, T., Soder, L., Aarne, O., Hammond, L., Ibrahim, L., Chan, A., Wills, P., Anderljung, M., Garfinkel, B., Heim, L., Trask, A., Mukobi, G., Schaeffer, R., Baker, M., Hooker, S., Solaiman, I., … Trager, R. (2025). *Open Problems in Technical AI Governance* (arXiv:2407.14981). arXiv. https://doi.org/10.48550/arXiv.2407.14981